\documentclass[10pt,twocolumn,letterpaper]{article}

\usepackage[pagenumbers]{iccv} 

\definecolor{iccvblue}{rgb}{0.21,0.49,0.74}
\usepackage[pagebackref,breaklinks,colorlinks,allcolors=iccvblue]{hyperref}

\def\paperID{8873} 
\def\confName{ICCV}
\def\confYear{2025}

\newcommand{\std}[1]{{\tiny$\pm$#1}}
\usepackage{xcolor}
\definecolor{mortarcolor}{RGB}{119,118,123}
\definecolor{brickcolor}{RGB}{107,70,71}
\newcommand\crule[3][black]{\textcolor{#1}{\rule{#2}{#3}}}
\usepackage{soul}
\usepackage{multirow}
\newcommand{\firstplace}[1]{\textcolor{red}{#1}}
\newcommand{\secondplace}[1]{\textcolor{blue}{#1}}
\usepackage{pifont}
\newcommand{\urlcode}[0]{\url{https://jmlipman.github.io/TopoMortar}}
\usepackage{float}
\usepackage[super]{nth}

\title{TopoMortar: A dataset to evaluate image segmentation methods focused on topology accuracy}

\author{Juan Miguel Valverde\textsuperscript{1,2}
\and
Motoya Koga\textsuperscript{3}
\and
Nijihiko Otsuka\textsuperscript{3}
\and
Anders Bjorholm Dahl\textsuperscript{1}
\and
\textsuperscript{1}Department of Applied Mathematics and Computer Science, Technical University of Denmark\\
\textsuperscript{2}A.I. Virtanen Institute, University of Eastern Finland\\
\textsuperscript{3}Department of Architecture, Faculty of Engineering, Sojo University, Japan}

\begin{document}
\maketitle
\begin{abstract}
We present TopoMortar, a brick wall dataset that is the first dataset specifically designed to evaluate topology-focused image segmentation methods, such as topology loss functions.
Motivated by the known sensitivity of methods to dataset challenges, such as small training sets, noisy labels, and out-of-distribution test-set images, TopoMortar is created to enable in two ways investigating methods' effectiveness at improving topology accuracy.
First, by eliminating dataset challenges that, as we show, impact the effectiveness of topology loss functions.
Second, by allowing to represent different dataset challenges in the same dataset, isolating methods' performance from dataset challenges.
TopoMortar includes three types of labels (accurate, pseudo-labels, and noisy labels), two fixed training sets (large and small), and in-distribution and out-of-distribution test-set images.
We compared eight loss functions on TopoMortar, and we found that clDice achieved the most topologically accurate segmentations, and that the relative advantageousness of the other loss functions depends on the experimental setting.
Additionally, we show that data augmentation and self-distillation can elevate Cross entropy Dice loss to surpass most topology loss functions, and that those simple methods can enhance topology loss functions as well.
TopoMortar and our code can be found at \urlcode.
\end{abstract}    
\section{Introduction}

Deep learning has demonstrated extraordinary potential for image segmentation, yet even state-of-the-art models \cite{kirillov2023segment} cannot guarantee the connectivity of thin tubular structures, such as axons, vessels, and fibers. As a result, minor misclassifications can break the continuity of these structures, compromising their subsequent quantification.
Topology loss functions \cite{hu2019topology} aim to address this issue by encouraging models to produce segmentations with the correct number of topological structures, such as connected components, holes, and hollows.
However, their effectiveness is not completely well understood due to limitations in the datasets used to evaluate them.

Topology loss functions have been evaluated on datasets with regions requiring precise connectivity, such as blood vessels.
Datasets, in addition to their high-level segmentation task (\eg, segmenting blood vessels on fundus retina images), present \textbf{challenges} that are rarely discussed or accounted for, such as class imbalance, small dataset size, noisy labels, pseudo-labels, and out-of-distribution (OOD) test-set images.
\begin{figure}
    \includegraphics[width=0.5\textwidth]{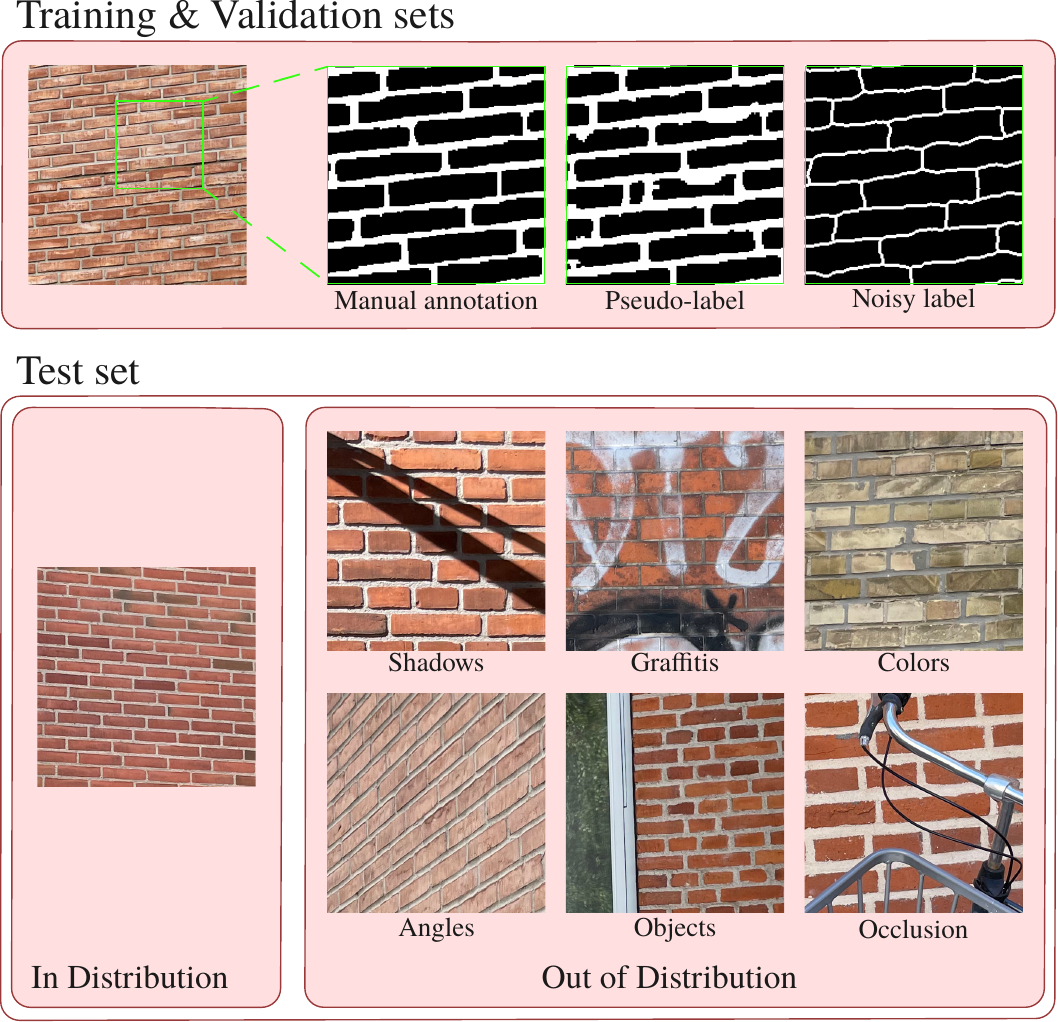}
   \caption{The TopoMortar dataset.} \label{fig:teaser}
\end{figure}
The entanglement between the dataset task and dataset challenges obscures understanding when and where topology loss functions improve topology accuracy.
For instance, a method addressing the same challenge across different datasets (\eg, Dice loss in class-imbalanced datasets) may increase topology accuracy simply because it improves accuracy by tackling that particular challenge; however, it will not improve topology accuracy in similar datasets with other challenges.
Separating dataset task from dataset challenges to investigate methods' \textbf{robustness} against such challenges allows elucidating whether methods learn data's topological properties \cite{el2021high}.
Additionally, topology loss functions have not yet been evaluated on a dataset without challenges (\eg, class imbalance, imperfect labels)---likely, because such a dataset does not exist.
A challenge-free dataset would reduce the possibilities for methods to increase topology accuracy by tackling dataset challenges.
Thus, an increase in topology accuracy can be attributed to the method's capability to learn the data's topological properties.
It is also unknown whether simple methods addressing dataset challenges, such as data augmentation and self-distillation, are more effective than topology losses, and whether topology loss functions can be further improved with such methods.
Our main contributions are:

\begin{itemize}

    \item We release the first dataset specifically acquired to investigate whether topology-focused methods improve topology accuracy and effectively learn data's topological properties. 
    
    \item We compare extensively, with 10 random seeds and statistical significance tests, Cross entropy Dice loss with six topology and one non-topology loss functions.

    \item We show that clDice was the only evaluated topology loss function that improved topology accuracy in most of the experiments, demonstrating its effectiveness.

    \item We demonstrate that data augmentation and self-distillation can increase topology accuracy even when optimizing topology loss functions.
    
\end{itemize}

\section{Related work} \label{sec:relatedwork}
Topology loss functions have been evaluated on many different image segmentation datasets where topology accuracy has been considered essential.
The most used datasets across 28 related studies \cite{hu2019topology,clough2019explicit,wang2020deep,li2020topology,clough2020topological,shit2021cldice,hu2021topology,wang2021single,araujo2021topological,yang2022topology,hu2022structure,wang2022ta,oner2021promoting,oner2023persistent,byrne2022persistent,ngoc2022topology,gupta2022learning,rouge2023cascaded,sofi2023image,he2023toposeg,lin2023dtu,gupta2024topology,hu2022learning,qi2023dynamic,liao2023segmentation,stucki2023topologically,shi2024centerline,kirchhoff2024skeleton} were DRIVE, Massachusetts Roads, and CREMI.
DRIVE \cite{staal2004ridge} is a dataset of fundus retina images for blood vessel segmentation that has 20 training and 20 test images.
The Massachusetts Roads dataset \cite{mnih2013machine} consists of 1171 aerial images for road segmentation split into 1108, 14, and 49 training, validation, and test set images.
The CREMI dataset\footnote{https://cremi.org} is comprised of three 3D electron-microscopy images of the brain tissue of adult Drosophila melanogaster.
Other datasets used to evaluate loss functions are CrackTree \cite{zou2012cracktree} (photographs of concrete cracks); ISBI12 \cite{arganda2015crowdsourcing} and ISBI13 \cite{arganda20133d} (electron-microscopy images of neurons); RoadTracer \cite{bastani2018roadtracer} and DeepGlobe \cite{demir2018deepglobe} (aerial images of roads); and ACDC \cite{bernard2018deep} and left ventricle UK biobank \cite{petersen2016uk} (cardiac magnetic resonance images).

These datasets present different challenges, making it difficult to determine whether improved topology accuracy stems from a method’s suitability to dataset challenges, its suitability to a specific task, or its effective learning of the data's topological properties.
The DRIVE dataset is extremely \textit{small}; around one-third of the training set images of the Massachusetts Roads dataset are \textit{corrupted} (see \Cref{app_sec:massachusetts_corrupted}); The CREMI dataset only provides the instance segmentation of the neurons, thus, each study had to derive its own neuron borders \textit{pseudo-labels}---a process that has not been documented and has been likely carried out differently (see \Cref{app_sec:cremi_pseudolabel}); CrackTree's labels are one-pixel width lines (\ie, \textit{noisy labels}, see \Cref{app_sec:cracktree_segmentation}).
On DRIVE---the most used dataset---previous studies have conducted 5-fold cross-validation on the 20 training set images \cite{kirchhoff2024skeleton}, 3-fold cross-validation on 30 images \cite{shit2021cldice}, 3-fold cross-validation on the 20 training set images \cite{hu2019topology,hu2021topology}, applied a 16-4-20 training, validation, and test set split \cite{shi2024centerline,gupta2024topology,araujo2021topological}, a 16-4 train-test split \cite{lin2023dtu}, or an unspecified split on the 20 training set images \cite{qi2023dynamic,hu2022structure}.
This inconsistency is likely due to a combination of factors, including its small size, the lack of a fixed train-validation split, and the unavailability of the test-set labels.
More details on the datasets and experimental discrepancies across studies can be found in \Cref{app_sec:fulldatasets}.
\section{TopoMortar}

Our dataset, TopoMortar, is a brick wall dataset consisting of 420 RGB images of 512 $\times$ 512 pixels for the task of mortar segmentation.
We have chosen brick walls with mortar because of the well-defined topological properties of the bricks and mortar.
The mortar allows variation in the labels, and the brick walls are well suited for testing specific shifts in domain such as occluding objects, color change, etc.
TopoMortar's task (\ie, mortar segmentation in red brick walls) is relatively simple by design; in contrast to existing datasets that are more complex, an improvement in topology accuracy in TopoMortar is less likely to be due to a more suitable choice of the neural network, optimizer, training time, etc.

TopoMortar is built to address previous dataset limitations and to avoid discrepancies in the experimental settings of future studies.
To this end, TopoMortar includes 1) a fixed training-validation-test set split (50-20-350), 2) two fixed training sets (50 and 10), 3) accurate, noisy, and pseudo-labels for the training and validation sets, 4) the manual annotations of all the images, and 5) several OOD test set images (85\% of the test set) divided into six groups portraying different challenges (see \cref{fig:teaser} (top)).
The training and validation sets contain images that align with the general concept of a red brick wall, \ie, reddish bricks with mortar horizontally and vertically oriented and without any shadows or objects.
The test set images are divided into seven groups: \textit{in-distribution} brick walls that are similar to the training and validation sets; brick walls with \textit{shadows} and \textit{graffitis}; brick walls with bricks of different \textit{colors}; brick walls images non-horizontally aligned taken from a different \textit{angle}; and brick walls with \textit{objects} in/next to them and objects \textit{occluding} the walls.
\Cref{fig:teaser} (bottom) shows an example of each category.
TopoMortar is larger than most datasets used in previous related studies (\Cref{app_sec:fulldatasets}).
Moreover, as our experiments demonstrate, its large training set consisting of 50 images suffices to achieve significantly higher topology accuracy than the small training set of only 10 images, allowing to study whether topology losses advantageousness decreases when increasing the training set size.

\textbf{TopoMortar allows to investigate whether methods are effective at improving topology accuracy} in two ways.
First, by \textbf{eliminating dataset challenges} (\ie, confounding factors such as scarce training data, inaccurate labels, and OOD test-set images) with TopoMortar's large training set, accurate labels, and ID test set.
Without dataset challenges, a method has limited ability to exploit specific dataset characteristics to increase accuracy and, therefore, topology accuracy.
Thus, an improvement in topology accuracy can be attributed to the effective learning of topological properties.
Second, by permitting to assess, on the same dataset, model \textbf{robustness against various dataset challenges}: scarce training data, pseudo-labels, noisy labels, and OOD test-set images.
By utilizing the same training set images (thus, fixing the dataset-related effects), an increase in topology accuracy across all challenges indicates the learning of topology information.
More details about TopoMortar, its labels, and its suitability for assessing topology accuracy can be found in \Cref{app_sec:topomortar_datasplit}.

\section{Experiments}
We conducted four sets of experiments.
First, we investigated the impact of dataset challenges and limitations on the effectiveness of topology loss functions in datasets used by previous work.
Second, we compared topology loss functions on TopoMortar in a setup without dataset challenges.
Third, we compared topology loss functions across different dataset challenges.
Fourth, we studied the extent to which two simple methods for tackling dataset challenges that do not directly aim at increasing topology accuracy (data augmentation and self-distillation) can actually increase topology accuracy.

We compared Cross entropy and Dice loss (CEDice), RegionWise loss \cite{valverde2023region}, TopoLoss \cite{hu2019topology}, TOPO \cite{oner2021promoting},  Warping loss \cite{hu2021topology}, clDice \cite{shit2021cldice}, Skeleton Recall \cite{kirchhoff2024skeleton}, and cbDice \cite{shi2024centerline}.
More details about the loss functions can be found in \Cref{app_sec:loss_functions}.
In addition to computing Dice coefficient \cite{dice1945measures} and Hausdorff distance (95th percentile) \cite{rote1991computing}, we computed the Betti errors.
The Betti 0 error ($\beta_0$) refers to the difference in the number of connected components, while the Betti 1 error ($\beta_1$) refers to the difference in the number of holes (in most cases corresponding to the bricks in TopoMortar).

All our experiments were run with 10 different random seeds, providing us with sufficient measurements to evaluate the significance of performance differences.
For this, we computed paired permutation tests with 10,000 random iterations.
We trained nnUNet for 12,000 iterations on batches of 10 images with deep supervision \cite{wang2015training} and stochastic gradient descent, with a learning rate of 0.01, nesterov momentum of 0.99, weight decay of $3 \times 10^{-5}$, and polynomial learning rate decay $(1 - \frac{iteration}{12000})^{0.9}$.
We employed 10 data augmentation transformations (see \Cref{app_sec:dataaugmentation}).
We implemented our experiments in MONAI \cite{cardoso2022monai} and PyTorch \cite{paszke2019pytorch}, and we ran our experiments in two clusters with several Tesla A100, V100, A10, and A40, ranging from 16 to 40 GB of GPU memory.

\subsection{Challenges and limitations in previous datasets} \label{sec:preliminaryexperiments}

\textbf{CREMI lacks true labels}, but for consistency with previous work we also utilized it.
\begin{table}
{\scriptsize
\begin{tabular}{|ll|ll|ll|}
\hline
 & & \multicolumn{2}{c|}{Betti error} & \multicolumn{2}{c|}{Dice}  \\
 & & D.A. & No D.A. & D.A. & No D.A. \\ \hline
\parbox[t]{1mm}{\multirow{8}{*}{\rotatebox[origin=c]{90}{\shortstack[c]{CREMI}}}} & CEDice & 2371\std{1046} & 1356\std{1126} & 0.81\std{0.0} & 0.75\std{0.0} \\
  & RWLoss & 3008\std{1008} & \textbf{1039\std{1145}} & 0.79\std{0.01} & 0.77\std{0.0} \\
  & TopoLoss & \textbf{1898\std{1090}} & 3393\std{1206} & 0.81\std{0.0} & 0.75\std{0.0} \\
  & TOPO & 16140\std{30223} & 20327\std{1687} & 0.63\std{0.22} & 0.78\std{0.0} \\
  & clDice & 2904\std{635} & 6421\std{1253} & 0.77\std{0.0} & 0.76\std{0.0} \\
  & Warping & 3795\std{998} & 1270\std{620} & 0.80\std{0.0} & 0.76\std{0.0} \\
  & SkelRecall & 2519\std{1323} & 2003\std{1198} & 0.75\std{0.01} & 0.75\std{0.0} \\
  & cbDice & 3407\std{951} & 2934\std{1460} & 0.81\std{0.0} & 0.76\std{0.0} \\
\hline \hline
 & & DRIVE & FIVES & DRIVE & FIVES \\ \hline
\parbox[t]{1mm}{\multirow{8}{*}{\rotatebox[origin=c]{90}{\shortstack[c]{Supervised}}}} & CEDice & 172.42\std{15.7} & 43.79\std{6.74} & 0.72\std{0.0} & 0.89\std{0.0} \\
  & RWLoss & \textbf{159.38\std{15.4}} & 163.9\std{17.78} & 0.74\std{0.0} & 0.76\std{0.01} \\
  & TopoLoss & \textbf{67.54\std{8.1}} & \textbf{42.51\std{7.15}} & 0.73\std{0.0} & 0.86\std{0.0} \\
  & TOPO & 221.6\std{37.08} & 76.83\std{16.51} & 0.72\std{0.01} & 0.81\std{0.01} \\
  & clDice & \textbf{61.04\std{8.89}} & \textbf{16.99\std{3.97}} & 0.73\std{0.0} & 0.86\std{0.0} \\
  & Warping & \textbf{115.62\std{11.38}} & 53.41\std{7.11} & 0.73\std{0.0} & 0.88\std{0.0} \\
  & SkelRecall & \textbf{128.86\std{15.28}} & \textbf{37.75\std{6.98}} & 0.72\std{0.0} & 0.83\std{0.01} \\
  & cbDice & \textbf{116.32\std{12.02}} & 55.69\std{7.77} & 0.73\std{0.0} & 0.87\std{0.0} \\
\hline \hline
 & & Supervised & Adele & Supervised & Adele \\ \hline
\parbox[t]{1mm}{\multirow{8}{*}{\rotatebox[origin=c]{90}{\shortstack[c]{CrackTree}}}} & CEDice & 84.3\std{9.66} & 51.97\std{7.56} & 0.78\std{0.01} & 0.75\std{0.01} \\
  & RWLoss & \multicolumn{1}{c}{-} & \multicolumn{1}{c|}{-} & \multicolumn{1}{c}{-} & \multicolumn{1}{c|}{-}  \\
  & TopoLoss & 145.82\std{18.18} & 134.87\std{19.59} & 0.52\std{0.02} & 0.47\std{0.02} \\
  & TOPO & \multicolumn{1}{c}{-} & \multicolumn{1}{c|}{-} & \multicolumn{1}{c}{-} & \multicolumn{1}{c|}{-}  \\
  & clDice & \textbf{22.13\std{28.54}} & \textbf{18.03\std{37.17}} & 0.54\std{0.12} & 0.34\std{0.08} \\
  & Warping & 94.85\std{10.71} & \textbf{46.37\std{7.35}} & 0.77\std{0.01} & 0.73\std{0.01} \\
  & SkelRecall & \textbf{14.03\std{5.82}} & \textbf{24.62\std{8.31}} & 0.32\std{0.01} & 0.3\std{0.01} \\
  & cbDice & 84.64\std{8.33} & 66.49\std{9.3} & 0.76\std{0.01} & 0.69\std{0.01} \\
\hline
\end{tabular}
} \caption{Mean and std. of Betti errors in previous datasets. Top: $\beta_1$ error on CREMI dataset, with vs. without data augmentation. Center: $\beta_0$ error in standard supervised training, DRIVE vs. FIVES datasets. Bottom: $\beta_0$ error on CrackTree dataset, standard supervised learning vs. Adele. \textbf{Bold}: Betti errors are lower and significantly different than CEDice loss.} \label{table:preliminary}
\end{table} 
\begin{table}
\centering
{\footnotesize
\begin{tabular}{|ll|llll|}
\hline
  & Loss & $\beta_0$ error & $\beta_1$ error & Dice & HD95  \\ \hline
\parbox[t]{1mm}{\multirow{8}{*}{\rotatebox[origin=c]{90}{\shortstack[c]{ID test set}}}} & CEDice & 3.31\std{2.62} & 2.13\std{1.02} & 0.91\std{0.00} & 1.87\std{0.03} \\
& RWLoss & 5.72\std{1.65} & 6.03\std{2.15} & 0.91\std{0.00} & 1.84\std{0.01} \\
& TopoLoss & 3.14\std{1.80} & 2.73\std{3.37} & 0.91\std{0.0} & 1.87\std{0.03} \\
& TOPO & 33.69\std{4.65} & 43.35\std{3.44} & 0.86\std{0.00} & 2.68\std{0.18} \\
& clDice & \textbf{1.17\std{0.54}} & \textbf{1.41\std{0.59}} & 0.91\std{0.00} & 1.83\std{0.01} \\
& Warping & 4.20\std{2.04} & 3.62\std{1.16} & 0.91\std{0.00} & 1.84\std{0.01} \\
& SkelRecall & 3.08\std{1.41} & \textbf{1.73\std{0.77}} & 0.91\std{0.00} & 1.90\std{0.03} \\
& cbDice & 4.24\std{2.08} & 3.81\std{2.28} & 0.91\std{0.00} & 1.85\std{0.02} \\
\hline
\parbox[t]{1mm}{\multirow{8}{*}{\rotatebox[origin=c]{90}{\shortstack[c]{OOD test set}}}} & CEDice & 197.25\std{24.21} & 87.81\std{21.32} & 0.65\std{0.01} & 37.65\std{1.20} \\
& RWLoss & 219.55\std{28.45} & 87.74\std{19.22} & 0.64\std{0.01} & 37.31\std{1.29} \\
& TopoLoss & \textbf{193.29\std{47.92}} & 99.33\std{28.05} & 0.65\std{0.01} & 37.33\std{1.06} \\
& TOPO & \textbf{117.73\std{19.37}} & 105.09\std{24.71} & 0.65\std{0.01} & 40.21\std{0.81} \\
& clDice & \textbf{81.33\std{13.16}} & \textbf{51.64\std{7.34}} & 0.66\std{0.01} & 36.90\std{1.29} \\
& Warping & 204.26\std{26.69} & \textbf{80.23\std{18.01}} & 0.65\std{0.01} & 37.07\std{1.25} \\
& SkelRecall & \textbf{180.23\std{24.44}} & 96.79\std{30.14} & 0.67\std{0.01} & 38.46\std{1.02} \\
& cbDice & 233.49\std{34.09} & 106.08\std{33.24} & 0.65\std{0.01} & 37.84\std{1.16} \\
\hline
\end{tabular}
}
\caption{Average performance (10 random seeds) on TopoMortar test set, separated into in-distribution (ID) and out-of-distribution (OOD) images. Training setup: Standard supervised learning, large training set, accurate labels. \textbf{Bold}: Betti errors are lower and significantly different than CEDice loss.}\label{table:standardbenchmark}
\end{table}
We divided CREMI into one image for training, one for validation, and one for testing, and we compared all loss functions, with and without data augmentation.
The only loss functions that achieved smaller and significantly different $\beta_1$ errors than CEDice were TopoLoss and RWLoss (see \Cref{table:preliminary}, ``CREMI").
However, since CREMI's pseudo-labels had many holes and the Betti errors only focused on their number, automatic segmentations with numerous holes, including incorrect ones, will show lower $\beta_1$ errors.
In consequence, a decrease in the $\beta_1$ error does not guarantee higher topology accuracy; instead, it might indicate that the segmentation had more incorrect holes (see \Cref{app_sec:cremi_segmentation}), making CREMI unsuitable for quantifying $\beta_1$ errors.
Data augmentation, which is heavily under-reported in the literature (\Cref{app_sec:fulldatasets}), was crucial to improving accuracy.
Furthermore, we observed that it was possible to make any loss function appear as the best by carefully selecting a random seed (\Cref{app_sec:importanceofseeds}).

\textbf{DRIVE is extremely small}, obscuring whether topology loss functions are particularly beneficial on fundus retina images or on datasets of small size.
To answer this question, we compared topology loss functions on DRIVE (13-2-5 train-validation-test split) alongside FIVES \cite{jin2022fives} (538-60-200 split), which is a similar but much larger dataset.
On the DRIVE dataset, six loss functions achieved smaller and significantly different $\beta_0$ errors than CEDice, whereas, on FIVES, only three of them achieved smaller and significantly different $\beta_0$ errors (\Cref{table:preliminary}, ``Supervised"), demonstrating the impact of scarce data on topology accuracy.
Loss functions generally performed better on the FIVES dataset, with CEDice gaining a nearly $\times$4 improvement in the $\beta_0$ error---the largest.
Additionally, while the Dice coefficients were similar across loss functions within the same dataset, the $\beta_0$ errors varied considerably.
clDice achieved the most topologically accurate segmentations on both datasets, and the relative effectiveness of the other loss functions varied.
For instance, Skeleton Recall was the \nth{5} most accurate loss on DRIVE, but the \nth{2} on FIVES.

\textbf{CrackTree's noisy labels} were annotated with one-pixel width lines (see \Cref{app_sec:cracktree_segmentation}).
We compared optimizing the topology loss functions via standard supervised learning, which is suboptimal for this type of labels, and via Adele \cite{liu2022adaptive}, which is a method designed for training deep learning models with noisy labels.
We divided CrackTree into a 147-17-42 train-validation-test split and we tackled class imbalance by multiplying the loss on each class by [0.2, 0.8].
All topology loss functions improved their topology accuracy when optimized via Adele (\Cref{table:preliminary}, ``CrackTree"), except Skeleton Recall loss that even with standard supervised learning it achieved the lowest $\beta_0$ errors.
RWLoss and TOPO led to empty masks.
Since the test-set labels are also noisy, the Dice coefficients hardly reflected segmentation quality.
For instance, clDice and Skeleton Recall, which achieved the lowest Dice coefficients, produced thick segmentations that corresponded better to the exact location of the concrete cracks than the ground truth (\Cref{app_sec:cracktree_segmentation}).
In general, Adele led all loss functions to produce thicker segmentations, decreasing their Dice coefficients.

\subsection{Benchmark on TopoMortar without challenges} \label{sec:standardbenchmark}
We evaluated topology loss functions on TopoMortar on a setup without dataset challenges, thus, \textbf{reducing the confounding factors} that could lead to an increase in topology accuracy without learning the data's topological properties.
To this end, we trained on TopoMortar's \textit{large} training set with \textit{accurate} labels, and we separated the performance measurements in the test set between \textit{ID} and OOD.
In other words, in this experiment, we accounted for no dataset challenges, as prior topology loss function studies, but, differently from those studies, we ensured our dataset had no such challenges, which, as we showed in \Cref{sec:preliminaryexperiments}, affected topology accuracy.
On top of the Betti errors, Dice coefficient, and HD95, we also measured local topology accuracy by computing the Betti error in a 128 $\times$ 128 sliding window.

clDice and Skeleton Recall were the only loss functions that achieved Betti errors lower and significantly different than CEDice on both the ID and OOD test-set images (see \Cref{table:standardbenchmark}).
In the ID test set, CEDice, TopoLoss, and Skeleton Recall produced segmentations with the second lowest $\beta_0$ errors, which, between them, were not significantly different.
In the OOD test set, TopoLoss and TOPO also achieved lower and significantly different $\beta_0$ errors than CEDice, while Warping did similarly on the $\beta_1$ error.
TOPO's low $\beta_0$ errors in the OOD dataset were due to over-segmentation, especially in the bricks with different colors (see \cref{app_sec:ood_segmentation_results}, ``colors").
The Dice coefficients and HD95 were similar across loss functions and, although they did not reflect segmentation quality too accurately, they signaled whether a loss function did not produce satisfactory segmentations (see TOPO in \Cref{table:standardbenchmark} (ID test set and Dice, and OOD test set and HD95)).
Since the local Betti errors were highly correlated to $\beta_0$ and $\beta_1$ errors (Pearson correlation $> 0.98$), we did not include them in the paper.

\subsection{Robustness to scarce training data, low-quality labels, and OOD images} \label{sec:exprobust}
We investigated on TopoMortar whether and to what degree existing topology loss functions enhance model \textbf{robustness} against scarce training data, inaccurate labels, and OOD images.
Studying model robustness by disentangling the different types of dataset challenges and other dataset-related factors allows elucidating if topology loss functions learn data's topological properties \cite{el2021high}.
First, we compared topology loss functions in a scarce training data setup (as in DRIVE) by using TopoMortar's \textit{small} training set.
Second, we compared them in a setup with labels generated semi- or fully automatically (as in CREMI) by using TopoMortar's \textit{pseudo-labels}.
Third, we compared topology loss functions in a setup with inaccurate labels resulting from a quick approximated human annotation (as in CrackTree) by using TopoMortar's \textit{noisy labels}.
Additionally, we separate the measurements distinguishing between ID and OOD test-set images.
In all the experiments, we trained the models via standard supervised learning and, unless otherwise specified, we employed TopoMortar's large training set and accurate labels.

\begin{table*}[ht]
    \centering
    \begin{minipage}[t]{0.43\textwidth}
{\scriptsize
\begin{tabular}{|ll|ll|ll|}
\hline
 &  & \multicolumn{2}{c|}{$\beta_0$ error}  & \multicolumn{2}{c|}{$\beta_1$ error}  \\
 & Test set $\rightarrow$ & ID & OOD & ID & OOD \\ \hline
\parbox[t]{1mm}{\multirow{8}{*}{\rotatebox[origin=c]{90}{\shortstack[c]{Small training set}}}} & CEDice & 9.6\std{4.4} & 182\std{16} & 4.6\std{3.3} & 61.5\std{8.6} \\
  & RWLoss & 12.4\std{7.2} & 201\std{17} & 6.4\std{2.8} & \textbf{56.5\std{12}} \\
  & TopoLoss & \secondplace{\textbf{8.7\std{3.0}}} & \secondplace{\textbf{152\std{14}}} & 4.8\std{5.1} & \textbf{58.1\std{8.9}} \\
  & TOPO & 60.0\std{20} & 312\std{34} & 43.3\std{3.5} & 109.6\std{30} \\
  & clDice & \firstplace{\textbf{6.2\std{6.2}}} & \firstplace{\textbf{127\std{15}}} & \firstplace{\textbf{4.0\std{3.0}}} & \firstplace{\textbf{38.7\std{8.0}}} \\
  & Warping & 11.4\std{6.5} & 190\std{20} & \secondplace{\textbf{4.5\std{2.8}}} & \secondplace{\textbf{43.6\std{12}}} \\
  & SkelRecall & 9.9\std{3.4} & \textbf{160\std{12}} & 5.2\std{4.0} & 69.2\std{10} \\
  & cbDice & 9.6\std{4.7} & \textbf{159\std{15}} & 4.9\std{2.9} & \textbf{54.8\std{8.3}} \\
\hline
\parbox[t]{1mm}{\multirow{8}{*}{\rotatebox[origin=c]{90}{\shortstack[c]{Pseudo-labels}}}} & CEDice & 12.3\std{1.5} & 126\std{26} & 12.0\std{1.5} & \secondplace{83.9\std{15}} \\
  & RWLoss & \textbf{11.4\std{1.4}} & \textbf{114.7\std{25}} & \textbf{11.11\std{2.0}} & 96.5\std{23} \\
  & TopoLoss & \textbf{10.4\std{1.1}} & \textbf{118\std{15}} & \secondplace{\textbf{10.1\std{1.4}}} & 110.3\std{33} \\
  & TOPO & 36.1\std{5.1} & 225.4\std{37} & 31.1\std{2.8} & 132\std{41} \\
  & clDice & \firstplace{\textbf{1.8\std{0.4}}} & \firstplace{\textbf{32.7\std{6.1}}} & \firstplace{\textbf{4.3\std{0.4}}} & \firstplace{\textbf{61.6\std{12}}} \\
  & Warping & 13.6\std{1.7} & 127\std{39} & 13.3\std{2.3} & 95.5\std{20} \\
  & SkelRecall & \secondplace{\textbf{10.4\std{1.6}}} & \secondplace{\textbf{113.6\std{25}}} & \textbf{10.4\std{1.7}} & 85.8\std{21} \\
  & cbDice & 13.1\std{3.5} & 141.4\std{33} & 14.7\std{2.5} & 123\std{28} \\
\hline
\parbox[t]{1mm}{\multirow{8}{*}{\rotatebox[origin=c]{90}{\shortstack[c]{Noisy labels}}}} & CEDice & 6.2\std{1.1} & 175\std{14} & \secondplace{5.1\std{0.6}} & 18.7\std{5.5} \\
  & RWLoss & 17.9\std{3.7} & 427.5\std{26} & 11.9\std{1.0} & \secondplace{\textbf{17.9\std{3.1}}} \\
  & TopoLoss & 7.2\std{2.5} & 205\std{12} & 6.6\std{0.4} & \firstplace{\textbf{14.9\std{1.3}}} \\
  & TOPO & 24.1\std{9.4} & \textbf{157\std{32}} & 10.9\std{4.8} & 140\std{22} \\
  & clDice & \secondplace{\textbf{5.0\std{0.5}}} & \secondplace{\textbf{126.5\std{13}}} & 6.8\std{0.5} & 48.2\std{16} \\
  & Warping & 8.5\std{1.7} & 285\std{19} & 9.5\std{0.6} & \textbf{18.0\std{1.8}} \\
  & SkelRecall & \firstplace{\textbf{3.8\std{1.6}}} & \firstplace{\textbf{113\std{14}}} & \firstplace{\textbf{1.5\std{0.4}}} & 44.6\std{7.9} \\
  & cbDice & 10.5\std{1.4} & 276\std{14} & 9.7\std{0.6} & 27.5\std{6.6} \\
\hline
\end{tabular}
} 
    \end{minipage}
    \begin{minipage}[t]{0.3\textwidth}
{\scriptsize
\begin{tabular}{|l|ll|ll|}
\hline
 & \multicolumn{2}{c|}{$\beta_0$ error}  & \multicolumn{2}{c|}{$\beta_1$ error}  \\
 & ID & OOD & ID & OOD \\ \hline
\parbox[t]{1mm}{\multirow{8}{*}{\rotatebox[origin=c]{90}{\shortstack[c]{D.A. (\textit{RandHue})}}}} & \secondplace{1.9\std{2.9}} & 69.7\std{15} & 2.0\std{5.0} & 53.3\std{12} \\
  & 3.3\std{1.0} & \textbf{62.8\std{5.8}} & 2.9\std{1.9} & \textbf{49.8\std{3.6}} \\
  & 2.9\std{4.2} & 85.3\std{20} & 3.3\std{7.8} & 80.1\std{19} \\
  & 19.7\std{2.3} & 97.2\std{11} & 40.6\std{1.4} & \textbf{49.9\std{11}} \\
  & \firstplace{\textbf{0.6\std{0.2}}} & \firstplace{\textbf{16.5\std{3.1}}} & \firstplace{\textbf{1.0\std{0.1}}} & \firstplace{\textbf{15.1\std{2.6}}} \\
  & 2.5\std{1.3} & \secondplace{\textbf{57.1\std{5.8}}} & \secondplace{1.9\std{1.3}} & \secondplace{\textbf{40.0\std{3.7}}} \\
  & 2.0\std{2.8} & 65.5\std{8.7} & 2.0\std{4.8} & 62.7\std{9.3} \\
  & 2.7\std{1.2} & 90.3\std{18} & 3.3\std{4.7} & 83.2\std{18} \\
\hline
\parbox[t]{1mm}{\multirow{8}{*}{\rotatebox[origin=c]{90}{\shortstack[c]{Pseudo + Self. dist.}}}} & \secondplace{3.8\std{0.4}} & 68.4\std{14} & 4.0\std{0.5} & 47.4\std{13} \\
   & 4.4\std{0.4} & 73.0\std{13} & 4.3\std{0.5} & \textbf{43.9\std{8.6}} \\
  & 7.6\std{1.4} & \secondplace{\textbf{34.9\std{6.6}}} & \secondplace{\textbf{3.6\std{0.4}}} & \textbf{31.7\std{6.7}} \\
  & 64.5\std{9.9} & 131.3\std{19} & 73.1\std{6.5} & 57.5\std{41} \\
  & \firstplace{\textbf{2.1\std{0.4}}} & \firstplace{\textbf{32.6\std{2.4}}} & \firstplace{\textbf{2.7\std{0.3}}} & \firstplace{55.9\std{10}} \\
  & 4.6\std{0.5} & \textbf{66.3\std{5.6}} & 4.8\std{0.5} & \secondplace{\textbf{42.7\std{7.4}}} \\
  & 4.7\std{0.4} & 81.5\std{17} & 5.4\std{0.6} & 61.8\std{15} \\
  & 4.5\std{0.6} & 78.4\std{14} & 4.8\std{0.5} & 61.4\std{19} \\
\hline
\parbox[t]{1mm}{\multirow{8}{*}{\rotatebox[origin=c]{90}{\shortstack[c]{Noisy + Self. dist.}}}} & 2.4\std{0.7} & 114\std{9.4} & 2.8\std{0.4} & \firstplace{13.5\std{1.7}} \\
  & 13.5\std{2.0} & 289\std{23} & 16.6\std{1.5} & 17.9\std{0.8} \\
  & 5.6\std{1.7} & 113\std{11} & 8.3\std{1.3} & \secondplace{15.7\std{0.7}} \\
  & 67.2\std{46} & 118\std{36} & 9.3\std{6.3} & 81.0\std{27} \\
  & \firstplace{\textbf{0.9\std{0.3}}} & \secondplace{\textbf{62.8\std{9.6}}} & \secondplace{\textbf{1.6\std{0.2}}} & 22.5\std{3.0} \\
  & 2.8\std{0.3} & 153\std{11} & 4.0\std{0.4} & 16.2\std{1.8} \\
  & \secondplace{\textbf{1.4\std{0.5}}} & \firstplace{\textbf{62.8\std{8.5}}} & \firstplace{\textbf{1.0\std{0.1}}} & 36.8\std{7.7} \\
  & 4.7\std{1.0} & 163\std{13} & 6.3\std{1.2} & 23.6\std{28} \\
\hline
\end{tabular}
}
    \end{minipage}
    \caption{Average performance on TopoMortar test set. Top: training setup as in \Cref{table:standardbenchmark} but on a small training set or with \textit{RandHue} data augmentation. Center: Pseudo-labels without, and with self-distillation. Bottom: Noisy labels, without and with self-distillation. \textbf{Bold}: Significantly lower than CEDice. \firstplace{Red}: Smallest average. \secondplace{Blue}: Second smallest.}

    \label{table:challenges}
\end{table*}

The models' performance decreased considerably after introducing the aforementioned dataset challenges, with an average Dice coefficient in the ID test set of 0.90, 0.86, and 0.68 in the scarce training data, pseudo-label, and noisy label setups, respectively (\Cref{app_sec:dicehd95}).
On both ID and OOD test sets, clDice achieved the lowest Betti errors when training on the small training set and when training on pseudo-labels, whereas Skeleton Recall loss was generally superior with noisy labels (\Cref{table:challenges} (left)).
The second-best topology loss function depended on the experimental setup and the Betti error (\Cref{table:challenges} (left), blue).
In the OOD test set, according to the $\beta_0$ error, TopoLoss was the second best in the small training set setup, Skeleton Recall in the pseudo-labels, and clDice in the noisy-labels experiment.
According to the $\beta_1$ error, Warping loss in the small training set, CEDice in the pseudo-labels, and RWLoss in the noisy-labels experiment.

\subsection{Topology losses with data augmentation and self-distillation} \label{sec:daself}
We studied the impact on topology accuracy of two simple methods for tackling dataset challenges.
First, we accounted for the presence of OOD images with a simple \textit{data augmentation} method that increased the colors' diversity in the images and that we applied with a probability of 50\%.
This data augmentation, which we refer to as \textit{RandHue}, converted the images to the HSV color space; randomly chose the same hue for all pixels; randomly shifted the saturation and value; and converted the image back to RGB (see examples in \Cref{app_sec:randhue_example}).
Second, we accounted for labels being pseudo-labels and noisy labels by training with \textit{self-distillation} \cite{hinton2015distilling}---a strategy known to be advantageous with those types of labels.
We employed self-distillation due to its simplicity and because it incorporated no extra hyper-parameters.
To keep the total iterations to 12,000 as in all our experiments, we trained the models for 4,000 iterations, generated soft labels for the training set, trained on those labels for another 4,000 iterations, and repeated the process one more time.

The segmentations and, particularly, their topology accuracy were generally better than in the previous experiment where no dataset challenge was directly tackled.
clDice also achieved the best segmentations in most cases, and Skeleton Recall also outperformed in the presence of noisy labels.
As in the previous experiment, the second-best topology loss function depended on the specific experimental setup.
Applying RandHue improved the Dice coefficients in the OOD images and decreased the Betti errors significantly in both the ID and OOD test set (\Cref{table:standardbenchmark} vs. \Cref{table:challenges} (top-right)).
The decrease in the $\beta_0$ error occurred across all OOD groups, whereas the decrease in the $\beta_1$ error occurred only in ``angles", ``colors", and ``shadows" (see \Cref{app_sec:topomortar_baselinerandhue_oodgroups}).
When training on pseudo-labels and noisy labels, self-distillation improved Dice coefficients and topology accuracy.
\section{Discussion}

We presented TopoMortar, a dataset specifically created to study the effectiveness of topology-focused image segmentation methods.
We showed that existing datasets exhibit challenges that were not previously considered and that influence topology accuracy.
We compared eight loss functions on TopoMortar on a setup without dataset challenges and then studied model robustness in the presence of those challenges.
Finally, we investigated the extent to which simple data augmentation and self-distillation can increase topology accuracy.

We evaluated topology loss functions on different datasets, following the standard experimental approach.
Additionally, we tackled the challenges of data scarcity and noisy labels in DRIVE and CrackTree datasets with a larger dataset and with a method to learn from noisy labels, respectively.
Our experiments revealed two key points.
First, no topology loss function was the best across all settings, which contrasts with topology loss function studies where the proposed loss function always outperforms the others.
Second, by tackling dataset challenges, not only topology accuracy improves but also the relative advantageousness of the topology loss functions changes.
Importantly, this experiment \textbf{does not reveal when and why} specific topology loss functions are advantageous, as comparing across datasets entangles dataset tasks and dataset challenges.
For instance, cbDice and Warping loss achieved higher topology accuracy than CEDice in the DRIVE dataset while, on CrackTree, they did not surpass CEDice.
This may indicate that cbDice and Warping loss lead to models robust against scarce training data but not against noisy labels; or that cbDice and Warping loss are particularly well suited for blood vessel segmentation; or both.

We evaluated the topology loss functions on TopoMortar \textbf{eliminating dataset challenges} by ensuring sufficient training data, training time, accurate labels, ID test-set images, strong data augmentation, and a state-of-the-art deep learning model.
This scenario was either assumed or not discussed in previous works.
In this challenging setup, only clDice and Skeleton Recall achieved $\beta_1$ errors smaller and significantly different than CEDice, with clDice also achieving smaller and significantly different $\beta_0$ errors, demonstrating the potential of skeletonization-based topology loss functions.

We investigated model \textbf{robustness against dataset challenges} after optimizing topology loss functions on TopoMortar.
In contrast to experiments on other datasets, TopoMortar allows fixing the dataset task (\ie, segmenting mortar in red brick walls), permitting to study the effect of each dataset challenge, individually, on the potential advantageousness of topology losses.
We observed 1) that clDice was generally the best-performing loss, 2) that Skeleton Recall worked best specifically under the presence of noisy labels, and 3) that the performance of the other losses varied depending on the dataset challenge.
These results, in line with the other experiments, indicate that clDice is truly effective at enhancing topology accuracy.
The outperformance of Skeleton Recall over clDice on noisy labels can be explained by its emphasis on the foreground region (true positives, false negatives), as, on TopoMortar's noisy labels, the foreground corresponds to a thicker and more accurate skeleton than what clDice produces.
Thus, it may be that with a different type of noisy labels \cite{algan2020label} Skeleton Recall performs differently.

We also studied the impact of data augmentation and self-distillation on topology accuracy.
These simple and well-established strategies improved the baseline CEDice when training on a large training set with accurate labels and when optimizing on noisy and pseudo-labels (\Cref{table:standardbenchmark,table:challenges}), making CEDice outperform the majority of topology loss functions.
Although it is unsurprising that data augmentation, self-distillation, and other methods \cite{bello2021revisiting} improve performance, limited research has investigated to what extent they can increase topology accuracy, or even if they can make standard models trained on CEDice surpass topology loss functions.
Considering that topology loss functions are generally computationally expensive CPU- and GPU-wise (\Cref{app:trainingresources}), our results demonstrate that focusing on improving regular accuracy by utilizing methods that account for dataset challenges can be a resource-friendly alternative to topology loss functions to increase topology accuracy.
Moreover, combining such methods with topology loss functions further improved topology accuracy in most cases, especially in the OOD test-set images (\Cref{table:challenges}).

TopoMortar was designed to be simple to prevent methods from exploiting dataset particularities to increase topology accuracy.
Despite mortar's relatively simple topology, topology accuracy on TopoMortar has a very high correlation with the topology accuracy on CREMI, DRIVE, FIVES, and CrackTree datasets (\Cref{app_sec:highcorrelation}), demonstrating the \textbf{generalizability} of results across biological, non-biological datasets, and structures with different topology.
\section{Conclusion}

Previous benchmarks on existing datasets have not allowed to completely understand whether methods improve topology accuracy merely by focusing on the dataset characteristics (thus, disregarding topology), or by learning data's topological properties.
In contrast, our TopoMortar dataset permits to study this research question by eliminating confounding factors and by enabling the investigation of model robustness against various dataset challenges.
clDice generally achieved the most topologically accurate segmentations while Skeleton Recall performed best on noisy labels.
Additionally, data augmentation and self-distillation helped in improving topology accuracy, even when employed jointly with topology loss functions.

\paragraph{Acknowledgements.}
This work was supported by Villum Foundation and NordForsk.
{
    \small
    \bibliographystyle{ieeenat_fullname}
    \bibliography{main}
}

\clearpage
\setcounter{page}{1}
\appendix
\setcounter{figure}{0} 
\setcounter{table}{0} 
\setcounter{equation}{0} 

\section{Massachusetts Roads corrupted images} \label{app_sec:massachusetts_corrupted}
\begin{figure}[h]
\centering
    \includegraphics[width=0.23\textwidth]{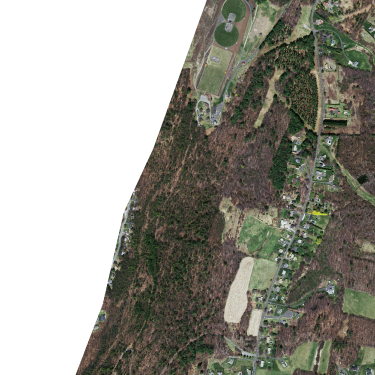}
    \includegraphics[width=0.23\textwidth]{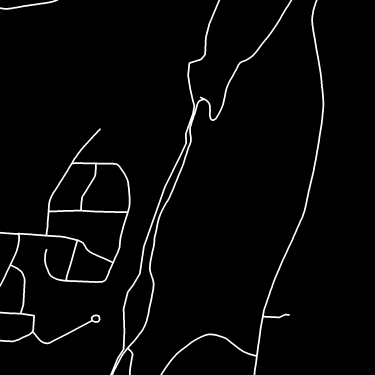}
    \includegraphics[width=0.23\textwidth]{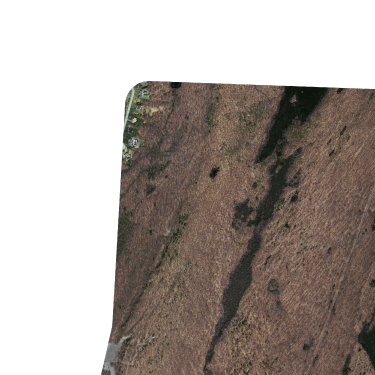}
    \includegraphics[width=0.23\textwidth]{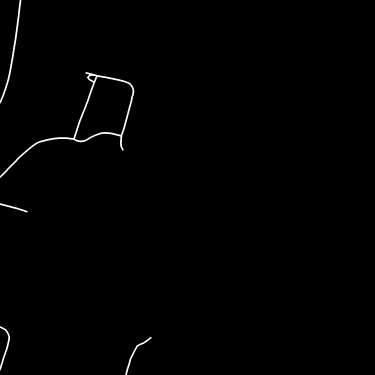}
   \caption{Two of the 320 corrupted images (left) with their ground truth (right).} \label{app_fig:massachussets_corrupted}
\end{figure}

The Massachusetts Roads dataset\footnote{https://www.cs.toronto.edu/~vmnih/data/}\footnote{https://www.kaggle.com/datasets/balraj98/massachusetts-roads-dataset} \cite{mnih2013machine}---one of the most popular datasets for evaluating topology loss functions---contains several images with large white patches that occlude the aerial images but not their ground truth.
Specifically, we counted 320 images in the training set (around one third of the total) that have over 10\% white pixels (\ie, [255, 255, 255]), indicating that they are corrupted.
\Cref{app_fig:massachussets_corrupted} shows two representative examples.
This issue, and whether it has been tackled and how, has been largely unreported.

\section{CREMI dataset} \label{app_sec:cremi_pseudolabel}

\begin{figure}[h]
\centering
    \includegraphics[width=0.48\textwidth]{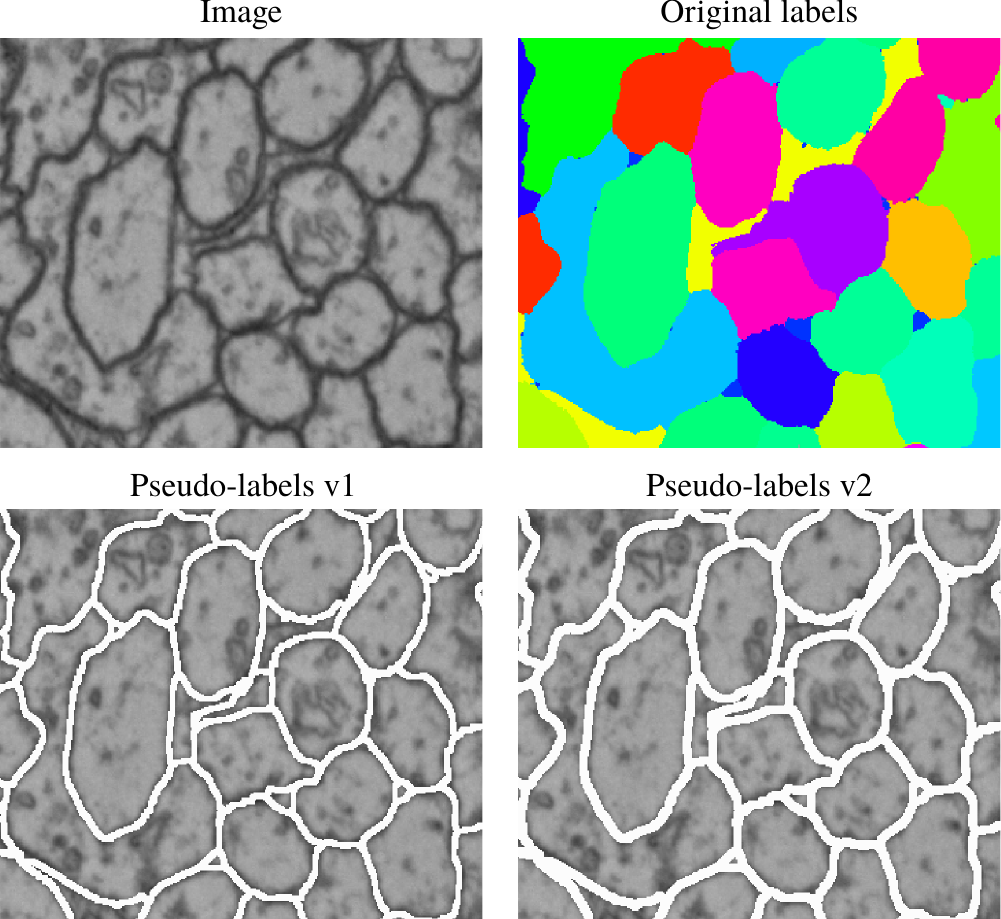}
   \caption{(a) Representative crop of CREMI dataset, (b) its ground-truth labels, and (c-d) two pseudo-labels derived with distance transform applying different thresholds.} \label{app_fig:cremi_pseudolabel}
\end{figure}

CREMI dataset is originally composed by three electron-microscopy images of the brain tissue of adult Drosophila melanogaster and the instance segmentation of the axons (see \Cref{app_fig:cremi_pseudolabel} (top)).
Previous studies focusing on topology loss functions have utilized this instance segmentation to derive pseudo-labels of the axon borders.
This process have not been exhaustively documented, and, as we report here, utilizing different thresholds on the distance maps can lead to pseudo-labels with very different size and topology.
For instance, applying a threshold value of ``4" (\Cref{app_fig:cremi_pseudolabel} bottom-right) increases by 33\% the size of the pseudo-labels compared to a threshold value of ``3" (\Cref{app_fig:cremi_pseudolabel} bottom-left), while the small cycles (\Cref{app_fig:cremi_pseudolabel} top-right dark blue) tend to disappear, thus, changing its topology.

\section{CrackTree segmentation example} \label{app_sec:cracktree_segmentation}
\begin{figure*}
\centering
    \includegraphics[width=\textwidth]{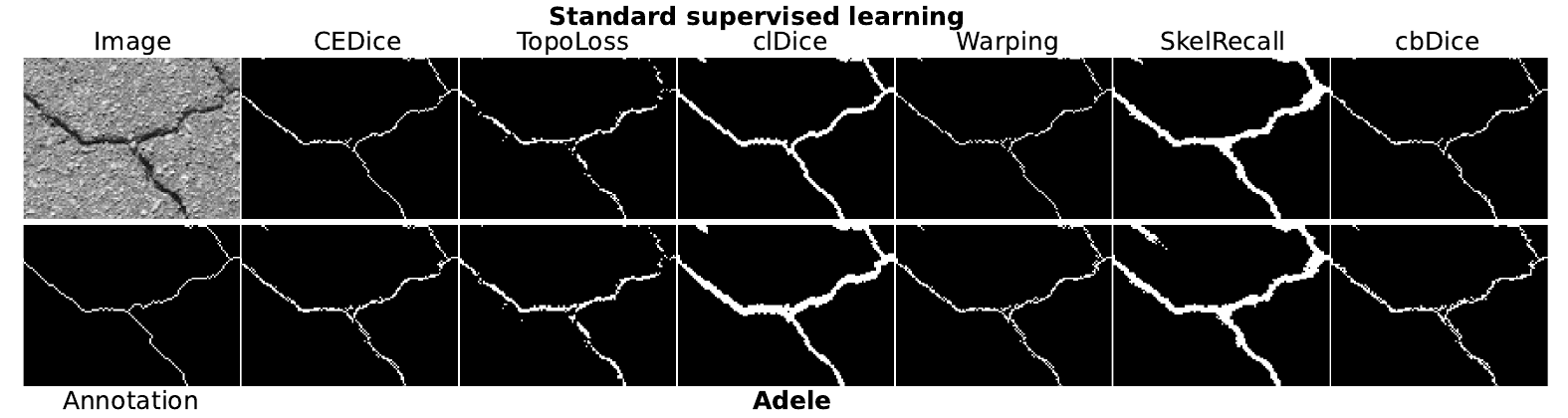}
   \caption{Representative segmentations in CrackTree. Top: Loss functions trained via standard supervised learning. Bottom: Trained via Adele.} \label{app_fig:cracktree_segmentation}
\end{figure*}

\Cref{app_fig:cracktree_segmentation} illustrates an example of CrackTree dataset, its corresponding annotation that is a line of only one-pixel width, and the segmentation with standard supervised learning and Adele \cite{liu2022adaptive}.

\section{Datasets used to evaluate topology loss functions}
\label{app_sec:fulldatasets}

\Cref{app_table:fulldatasets} lists the datasets used by, at least, two of the 28 recent studies that we examined that proposed a topology-focused image segmentation method.
\Cref{app_table:fulldatasets} also shows the training and optimization settings of these works, where we can observe a large discrepancy across experiments in previous works.
In addition to these datasets, 35 other datasets were used by only one study.

\begin{table*}[]
{\fontsize{9}{5}\selectfont
\begin{tabular}{lllllll}
\textbf{Dataset}                                                  & \textbf{Information}              & \textbf{Training configuration}              & \textbf{Architecture}      & \textbf{Runs} & \textbf{D.A.}                        & \textbf{Study}                                           \\
\hline
DRIVE \cite{staal2004ridge}             & - \textbf{40 2D images}           & 5-fold xval on 20 images            & nnUNet, HRNet     & ?    & ?                         & \cite{kirchhoff2024skeleton}   \\
                                                         & - Blood vessels          & 3-fold xval on 30 images            & UNet, FCN         & ?    & ?                         & \cite{shit2021cldice}          \\
                                                         & - Optical coherence      & Unspecified split on 20 images      & UNet              & ?    & ?                         & \cite{hu2022structure}         \\
                                                         & tomography               & 3-fold xval on 20 images            & ?                 & ?    & ?                         & \cite{hu2019topology}          \\
                                                         &                          & 16-4-20 train-val-test              & nnUNet            & ?    & \checkmark & \cite{shi2024centerline}       \\
                                                         &                          & 16-4-20 train-val-test              & UNet              & ?    & ?                         & \cite{gupta2024topology}       \\
                                                         &                          & 3-fold xval                         & ProbabilisticUnet & ?    & ?                         & \cite{hu2022learning}          \\
                                                         &                          & 16-4 train-test                     & UNet              & ?    & ?                         & \cite{lin2023dtu}              \\
                                                         &                          & 3-fold xval on 20 images            & UNet              & ?    & ?                         & \cite{hu2021topology}          \\
                                                         &                          & 16-4-20 train-val-test              & UNet              & 2    & \checkmark & \cite{araujo2021topological}   \\
                                                         &                          & Unspecified split on 20 images      & DSCNet            & ?    & \checkmark & \cite{qi2023dynamic}           \\
                                                         &                          & 20-20 train-test                    & Own method        & ?    & \checkmark & \cite{liao2023segmentation}    \\
                                                         \hline
Massachusetts                                            & - \textbf{1171 2D images}         & Predefined split on 804 images      & nnUNet, HRNet     & ?    & ?                         & \cite{kirchhoff2024skeleton}   \\
Roads \cite{mnih2013machine}            & - Roads                  & 3-fold xval on 120 images           & UNet, custom FCN  & ?    & ?                         & \cite{shit2021cldice}          \\
                                                         & - Satellite imagery      & 3-fold xval                         & UNet              & ?    & ?                         & \cite{hu2022structure}         \\
                                                         &                          & 3-fold xval on 1108 images          & ?                 & ?    & ?                         & \cite{hu2019topology}          \\
                                                         &                          & 100-24 train-test                   & UNet              & ?    & ?                         & \cite{stucki2023topologically} \\
                                                         &                          & 3-fold xval                         & UNet              & ?    & \checkmark & \cite{oner2021promoting}       \\
                                                         &                          & 1108-14-49 train-val-test           & UNet              & ?    & ?                         & \cite{gupta2024topology}       \\
                                                         &                          & 3-fold xval on 1108 images          & UNet              & ?    & ?                         & \cite{hu2021topology}          \\
                                                         &                          & 3-fold xval                         & UNet              & 3    & \checkmark & \cite{oner2023persistent}      \\
                                                         &                          & 1108-14-49 train-val-test           & DSCNet            & ?    & \checkmark & \cite{qi2023dynamic}           \\
                                                         &                          & 10-1098 train-test                  & Own method        & ?    & \checkmark & \cite{liao2023segmentation}    \\
                                                         \hline
CREMI\footnote{https://cremi.org}       & - \textbf{3 3D images}            & 3-fold xval on 324 slices           & UNet              & ?    & ?                         & \cite{shit2021cldice}          \\
                                                         & - Neuron borders         & 3-fold xval                         & UNet              & ?    & ?                         & \cite{hu2022structure}         \\
                                                         & - Electron microscopy    & 3-fold xval on 125 slices           & Unspecified       & ?    & ?                         & \cite{hu2019topology}          \\
                                                         &                          & 100-25 slices (train-test)          & UNet              & ?    & ?                         & \cite{stucki2023topologically} \\
                                                         &                          & 3-fold xval on 125 slices           & ProbabilisticUnet & ?    & ?                         & \cite{hu2022learning}          \\
                                                         &                          & 3-fold xval on 125 slices           & UNet              & ?    & ?                         & \cite{hu2021topology}          \\
                                                         &                          & 3-fold xval                         & ConvLSTM          & ?    & \checkmark & \cite{yang2022topology}        \\
                                                         \hline
ISBI12 \cite{arganda2015crowdsourcing}  & - \textbf{30 2D slices}           & 3-fold xval                         & ?                 & ?    & ?               & \cite{hu2019topology}          \\
                                                         & - Neuron borders         & 3-fold xval                         & UNet              & ?    & ?               & \cite{hu2021topology}          \\
                                                         & - Electron microscopy    & 3-fold xval                         & ConvLSTM          & ?    & \checkmark & \cite{yang2022topology}        \\
                                                         &                          & 24-6 train-test                     & UNet              & ?    & \checkmark & \cite{sofi2023image}           \\
                                                         \hline
ISBI13 \cite{arganda20133d}             & - \textbf{100 2D slices}          & 3-fold xval                         & ?                 & ?    & ?                         & \cite{hu2019topology}          \\
                                                         & - Neuron borders         & 3-fold xval                         & ProbabilisticUnet & ?    & ?                         & \cite{hu2022learning}          \\
                                                         & - Electron microscopy    & 3-fold xval                         & UNet              & ?    & ?                         & \cite{hu2021topology}          \\
                                                         &                          & 3-fold xval                         & ConvLSTM          & ?    & \checkmark & \cite{yang2022topology}        \\
                                                         \hline
RoadTracer \cite{bastani2018roadtracer} & - \textbf{300 2D images}          & 180-120 train-val                   & UNet              & ?    & ?                         & \cite{hu2022structure}         \\
                                                         & of 40 cities             &                                     &                   &      &                           &                                                 \\
                                                         & - Roads                  & 25-15 cities train-val              & UNet              & ?    & \checkmark & \cite{oner2021promoting}       \\
                                                         & - Satellite imagery      & 25-15 cities train-val              & UNet              & 1    & \checkmark & \cite{oner2023persistent}      \\
                                                         \hline
CrackTree \cite{zou2012cracktree}       & - \textbf{206 2D images}          & 3-fold xval                         & ?                 & ?    & ?                         & \cite{hu2019topology}          \\
                                                         & - Concrete cracks        & 3-fold xval                         & UNet              & ?    & ?                         & \cite{hu2021topology}          \\
                                                         & - Photographs            &                                     &                   &      &                           &                                                 \\
                                                         \hline
DeepGlobe \cite{demir2018deepglobe}     & - \textbf{8570 2D images}         & 4696-1530 train-val                 & UNet              & ?    & ?                         & \cite{hu2022structure}         \\
                                                         & - Roads                  & 4696-1530 train-val                 & UNet              & ?    & \checkmark & \cite{oner2021promoting}       \\
                                                         & - Satellite imagery      &                                     &                   &      &                           &                                                 \\
                                                         \hline
TopCow \cite{yang2023benchmarking}      & - \textbf{110+90 3D images}       & Predefined train-test               & nnUNet, HRNet     & ?    & ?                         & \cite{kirchhoff2024skeleton}   \\
                                                         & - Circle of Willis       & 72-18 (CTA) train-val               & nnUNet            & ?    & \checkmark & \cite{shi2024centerline}       \\
                                                         & - 110 MRI, 90 CTA        &                                     &                   &      &                           &                                                 \\
                                                         \hline
Parse2022 \cite{luo2023efficient}       & - \textbf{100 3D images}          & 80-20 train-test                    & nnUNet            & ?    & \checkmark & \cite{shi2024centerline}       \\
                                                         & - Pulmonary arteries     & 4-fold xval                         & UNet              & ?    & ?                         & \cite{gupta2024topology}       \\
                                                         & - CT                     &                                     &                   &      &                           &                                                 \\
                                                         \hline
Left ventricle                                           & - \textbf{900 images}             & Various settings                    & UNet              & 20   & ?                         & \cite{clough2019explicit}      \\
UK Biobank \cite{petersen2016uk}        & - Ventricles             & Various settings                    & UNet              & ?    & ?                         & \cite{clough2020topological}   \\
                                                         & - Cardiac MRI            &                                     &                   &      &                           &                                                 \\
                                                         \hline
ACDC \cite{bernard2018deep}             & - \textbf{150 patients (4D)}      & 100-50 train-test                   & UNet              & ?    & ?                         & \cite{clough2020topological}   \\
                                                         & - Ventricles, Myocardium & 300-150-150 (slices) train-val-test & UNet              & ?    & \checkmark & \cite{byrne2022persistent}     \\
                                                        
                                                         & - Cardiac MRI            &                                     &                   &      &                           &   \\                             \hline                
\end{tabular}
}
\caption{Datasets and experimental setting across studies on topology and image segmentation. Information: \textbf{Number of images}, target region of interest, and imaging modality. Runs: Number of independent runs with different random seeds. D.A.: Data augmentation. ?: Unspecified.}
\label{app_table:fulldatasets}
\end{table*}

\newpage

\section{TopoMortar dataset details}
\label{app_sec:topomortar_datasplit}

\begin{figure}
\centering
    \includegraphics[width=0.47\textwidth]{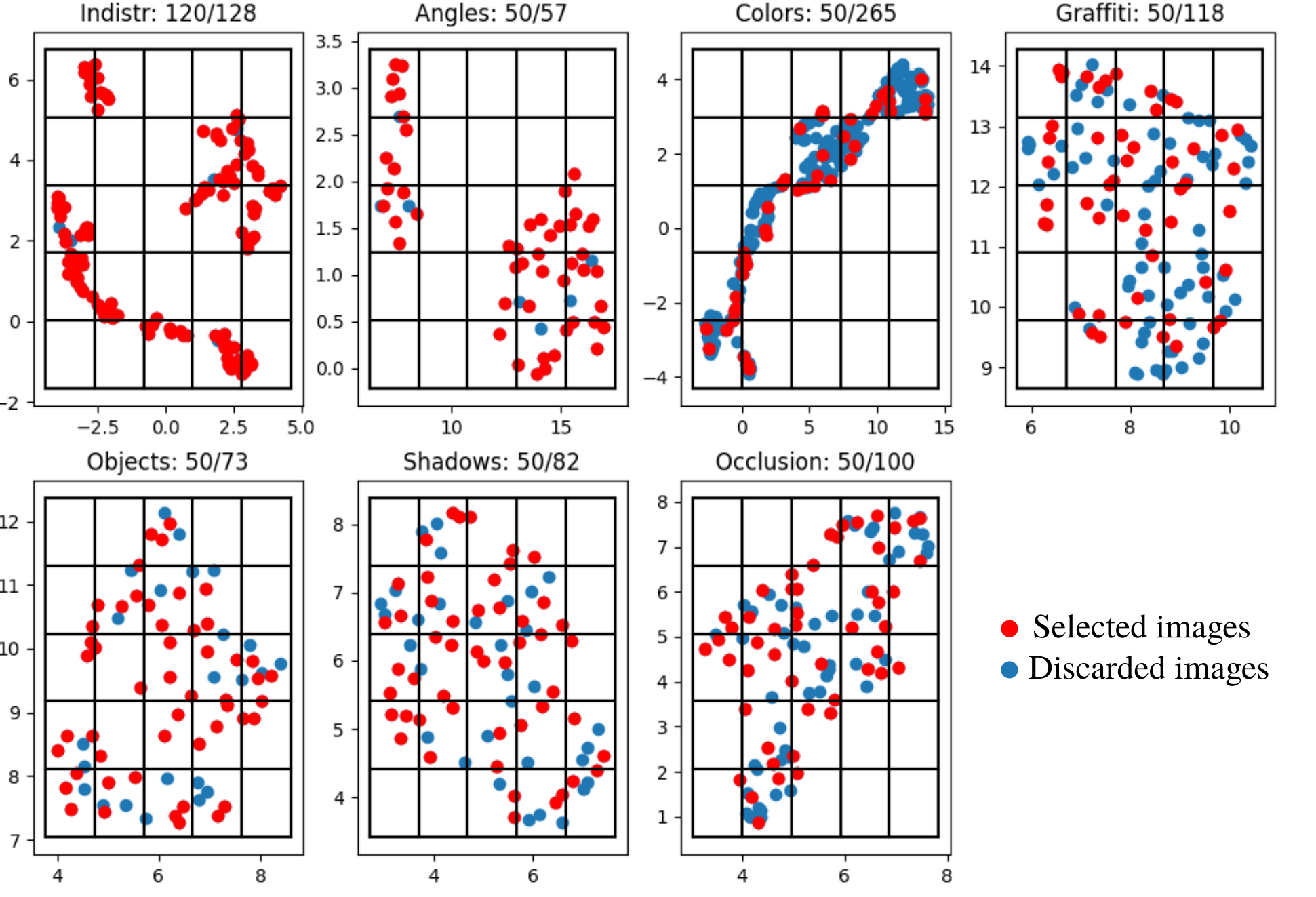}
   \caption{UMAP embeddings of TopoMortar's cropped patches separated by category.} \label{app_fig:topomortar_datasetsplit_umap}
\end{figure}

\paragraph{Data acquisition, processing, and split}
We took 195 photographs of brick walls and we manually cropped them into 823 512 $\times$ 512 non-overlapping patches that were, subsequently, divided into in-distribution and the six out-of-distribution categories (shadows, graffitis, colors, angles, objects, occlusion).
For the purpose of creating the dataset split, we down-scaled the patches to 256 $\times$ 256, flatten them, and, for each patch, we computed the histogram (1000 bins) of its intensity values, resulting into 823 1000-length vectors.
We then grouped the vectors by their category and reduced their dimensionality to two components with UMAP \cite{mcinnes2018umap}.
We divided the embedded space into a 5 $\times$ 5 grid, and we uniformly sampled the images from the cells, achieving the desired number of images per category: 120 images for in-distribution, and 50 for each of the six out-of-distribution groups.
\Cref{app_fig:topomortar_datasetsplit_umap} illustrates this process.
Finally, we randomly divided the in-distribution group into 50-20-50 for the training, validation, and test set, and included all the out-of-distribution images in the test set.

\paragraph{Labels}
We obtained the \textit{pseudo-labels} for the in-distribution (ID) images by fitting the images into a Gaussian Mixture model of two components (mortar and brick).
Since the mortar and bricks are grayish and reddish in the majority of the images, we initialized the model with means $\mu_{mortar} = [119, 118, 123]$ \crule[mortarcolor]{0.3cm}{0.3cm}, $\mu_{brick} = [107, 70, 71]$ \crule[brickcolor]{0.3cm}{0.3cm}.
We then removed the connected components smaller than 300 pixels, and applied binary dilation followed by binary erosion.
The manual annotations (\textit{accurate} labels) for the ID images were obtained by carefully refining the pseudo-labels manually. 
For the out-of-distribution (OOD) images, a few models trained on TopoMortar's training set were ensembled and the predictions were manually corrected.
The manual annotation process took approximately 210 hours.
Finally, we generated the \textit{noisy labels} by skeletonizing the manual annotations and applying random elastic deformations and binary dilation, imitating rapid manual annotations with small human errors (see \Cref{fig:teaser}).

\begin{figure}
\centering
    \includegraphics[width=0.47\textwidth]{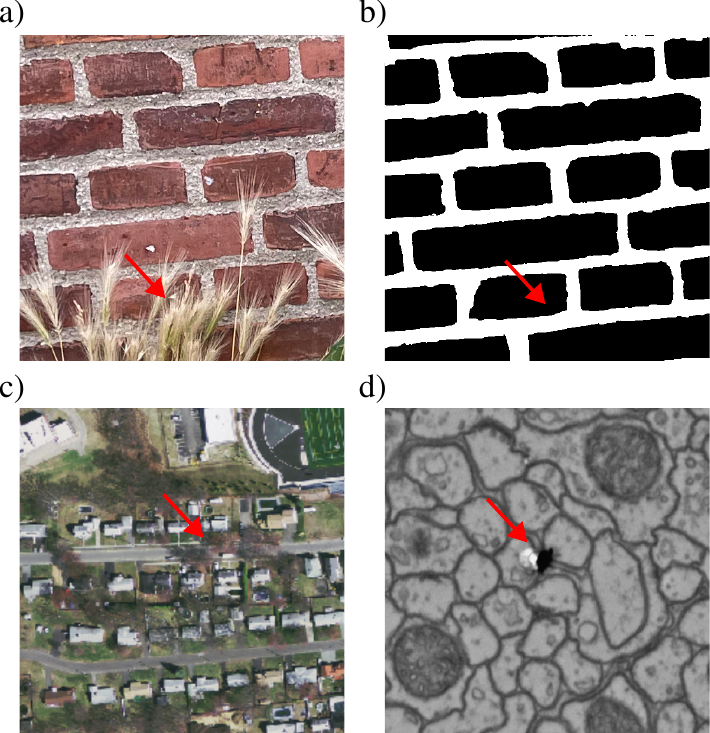}
   \caption{Structures that appear disconnected due to occlusion. a-b) TopoMortar's ``occlusion" image and its ground truth. c) Massachusetts Roads' image with trees occluding the road. d) CREMI crop with debris occluding the myelin.}\label{app_fig:unconnected}
\end{figure}

\paragraph{Suitability for assessing topology accuracy}
TopoMortar's OOD test set was deliberately designed to be difficult given the training set images, as it includes unseen scenarios with local and global differences in the intensity values (shadows, graffitis, colors), different brick orientations (angles), and the appearance of objects within, near, and occluding the brick walls (objects, occlusion).
The \textit{occlusion} category allows assessing whether the evaluated methods can connect structures that appear unconnected and, we know, are actually connected.
This is particularly relevant for amodal segmentation, where the goal is to predict the complete structure even when parts are occluded or hidden.
Thus, this category helps to elucidate whether the model has gained information about the true topology of the structures, which is essential for segmenting, \eg, roads in aerial images occluded by trees, myelin in electron-microscopy images with debris, and structures in medical images with limited resolution (see \Cref{app_fig:unconnected}).
Additionally, the comparatively small size of TopoMortar's images (512 $\times$ 512 pixels) lessens GPU memory requirements, thereby offering ample capacity for methods with high memory demands.
Moreover, unlike in most datasets that focus on either 0-dimensional topological structures (connected components) or 1-dimensional topological structures (holes), in TopoMortar both topological structures are relevant, as they correspond to the mortar and bricks, respectively.

\section{Loss functions} \label{app_sec:loss_functions}
 We compared eight loss functions, including six topology loss functions, with different characteristics.
\textit{Non-topology loss functions}: The combination of Cross entropy and Dice loss (CEDice), which are the most utilized loss functions in image segmentation; RegionWise loss \cite{valverde2023region}, that is based on distances to the structures' borders and has been shown to improve topology accuracy \cite{liu2024enhancing}. 
\textit{Persistence-homology-based loss functions}: TopoLoss \cite{hu2019topology}, that finds via persistence homology \cite{edelsbrunner2002topological} the pixels that lead to topological errors.
\textit{Distance-maps-based topology loss functions}: TOPO \cite{oner2021promoting} and Warping loss \cite{hu2021topology}, that employ distance maps to identify the critical areas that change the topology of the segmentations.
\textit{Skeletonization-based loss functions}: clDice \cite{shit2021cldice}, Skeleton Recall \cite{kirchhoff2024skeleton}, and cbDice \cite{shi2024centerline}, that focus on the accuracy of the segmentations' skeletons.
In our experiments, we utilized the official Github source code of those seven loss functions.

\paragraph{RegionWise loss}
In the original study \cite{valverde2023region}, the region-wise maps $\mathbf{z}$ corresponded to the distance to the border of the ground truth.
Since, in the second and third self-distillation iterations, the pseudo-labels are softmax probabilities, we computed region-wise loss differently.
We considered the softmax probabilities as if they were distances, and the probabilities $>0.9$ were considered to indicate the presence of the foreground.

\paragraph{TopoLoss}
In agreement with the original study \cite{hu2019topology}, we combined TopoLoss with Cross entropy loss (\ie, $\mathcal{L} = \mathcal{L}_{ce} + \lambda \mathcal{L}_{warp}$).
Due to the long time required to compute TopoLoss, we set $\lambda=0$ during the first 70\% of the training time, and $\lambda=100$ during the remaining 30\%.
Additionally, we set $path\_size=50$.

\paragraph{TOPO}
In agreement with the original study \cite{oner2021promoting}, we combined TOPO windowed loss with Mean square error loss (\ie, $\mathcal{L} = \mathcal{L}_{MSE} + \alpha \mathcal{L}_{TOPO}$).
We set $\alpha=0.001$.
Additionally, since models trained with TOPO windowed loss produced outputs of only one channel, in the self-distillation experiments pseudo-labels were binarized.

\paragraph{clDice loss}
In agreement with the original study \cite{shit2021cldice}, we combined clDice loss with Dice loss (\ie, $\mathcal{L} = (1-\alpha)(1-\mathcal{L}_{dice}) + \alpha(1-\mathcal{L}_{clDice})$).
The hyper-parameters that we used were: $\alpha=0.5$, $k=3$ (number of iterations).

\paragraph{Warping loss}
In agreement with the original study \cite{hu2022structure}, we combined Warping loss with Dice loss (\ie, $\mathcal{L} = \mathcal{L}_{dice} + \lambda \mathcal{L}_{warp}$).
Due to the long time required to compute Warping loss, we set $\lambda=0$ during the first 70\% of the training time, and $\lambda=0.1$ during the remaining 30\%.

\paragraph{Skeleton Recall loss}
In agreement with the original study \cite{kirchhoff2024skeleton}, we combined Skeleton Recall loss with Cross entropy loss (\ie, $\mathcal{L} = \mathcal{L}_{ce} + \lambda \mathcal{L}_{skel\_recall}$).
We set $\lambda=1$.

\paragraph{cbDice loss}
In agreement with the original study \cite{shi2024centerline}, we combined Centerline boundary Dice loss with Cross entropy and Dice loss (\ie, $\mathcal{L} = 0.5 \mathcal{L}_{ce} + \frac{\alpha}{2(\alpha+\beta)}\mathcal{L}_{dice} + \frac{\beta}{2(\alpha+\beta)}\mathcal{L}_{cbDice}$).
We set $\alpha=\beta=1$.

\section{Data Augmentation} \label{app_sec:dataaugmentation}

\Cref{app_table:dataaugmentation} lists the data augmentation transformations employed in all our experiments.

\begin{table}[h]
\centering
{\footnotesize
\begin{tabular}{ll}
\textbf{Transformation (probability)}         & \textbf{Parameters}                          \\
\hline
Rand. rotation (0.2)                         & {[}-$\pi$, $\pi${]}                                \\
Rand. scale (0.2)                            & {[}0.7, 1.4{]}                               \\
Gaussian noise (0.1)                          & N(0, 0.1)                                    \\
Gaussian blur (0.2)                           & $\sigma_x$={[}0.5, 1{]}, $\sigma_y$={[}0.5, 1{]} \\
Rand. intensity scale (0.15)                 & {[}-0.25, 0.25{]}                            \\
Rand. intensity scale (fixed mean) (0.15) & {[}-0.25, 0.25{]}                            \\
Rand. low resolution (0.25)                  & {[}0.5, 1{]}                                 \\
Rand. adjust contrast (inverted image) (0.1) & {[}0.7, 1.5{]}                               \\
Rand. adjust contrast (0.1)                  & {[}0.7, 1.5{]}                               \\
Rand. axis flip (0.5)                        & -                                           
\end{tabular}
}
\caption{Data augmentation used in all our experiments.}
\label{app_table:dataaugmentation}
\end{table}

\section{CREMI segmentation results} \label{app_sec:cremi_segmentation}
\begin{figure}[h]
\centering
    \includegraphics[width=0.48\textwidth]{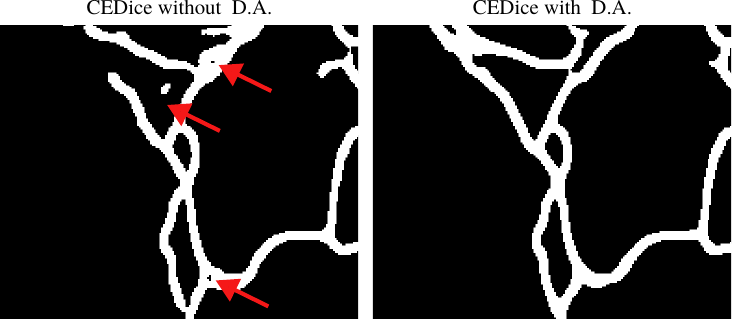}
   \caption{A slice of CREMI dataset segmented by CEDice with and without data augmentation. Left: The lack of data augmentation led to more holes, but also to more wrong openings. Right: Data augmentation helped in achieving more realistic segmentations.} \label{app_fig:cremi_wrongholes}
\end{figure}

In CREMI, 1-dimensional topological structures (\ie, holes, cycles) correspond to axons.
However, the $\beta_1$ error, which is the difference in the number of holes between the ground truth and the automatic prediction, cannot distinguish between correct and incorrect holes.
Since applying no data augmentation leads to inaccurate segmentations with more incorrect holes in the borders and CREMI's pseudo-labels contain numerous holes, it appears that the lack of data augmentation leads to topologically more correct segmentations.
In other words, segmentations with more wrong holes often achieved smaller $\beta_1$ errors (\Cref{app_fig:cremi_wrongholes}), making CREMI unsuitable to measure $\beta_1$ errors.

\section{Any loss can be made appear the best} \label{app_sec:importanceofseeds}
\begin{figure}[h]
\centering
    \includegraphics[width=0.48\textwidth]{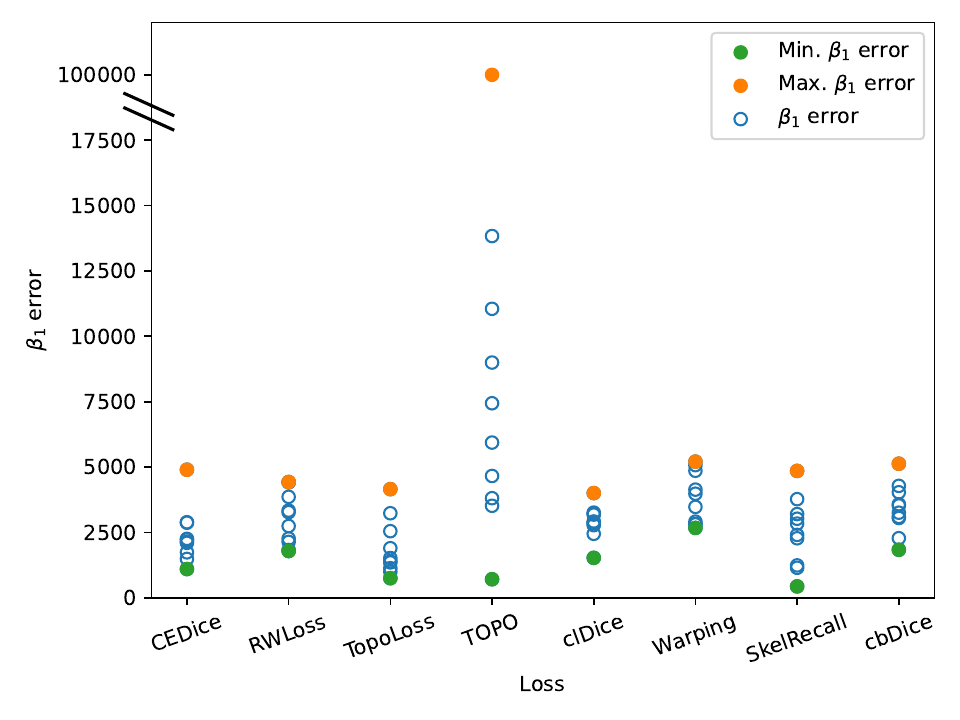}
   \caption{$\beta_1$ errors for each loss function in CREMI dataset (with data augmentation). Green: Smallest $\beta_1$ error. Orange: Largest $\beta_1$ error. Blue: Others.} \label{app_fig:importanceofseeds}
\end{figure}

We observed that the majority of previous works did not report the use of more than one random seed---possibly due to the large computational requirements associated to topology loss functions.
In this study, where we run every experiment with 10 random seeds, we noticed that the large variability in the Betti errors permits to portray almost any loss function as the most topologically accurate by carefully selecting a random seed.
\Cref{app_fig:importanceofseeds} illustrates this issue in CREMI dataset (with data augmentation): to make any loss appear the best, one would need to select the random seed corresponding to the green circle, and for the others the random seed corresponding to the orange circle.

\section{Segmentations on TopoMortar's OOD test set} \label{app_sec:ood_segmentation_results}
\begin{figure*}[h]
\centering
    \includegraphics[width=\textwidth]{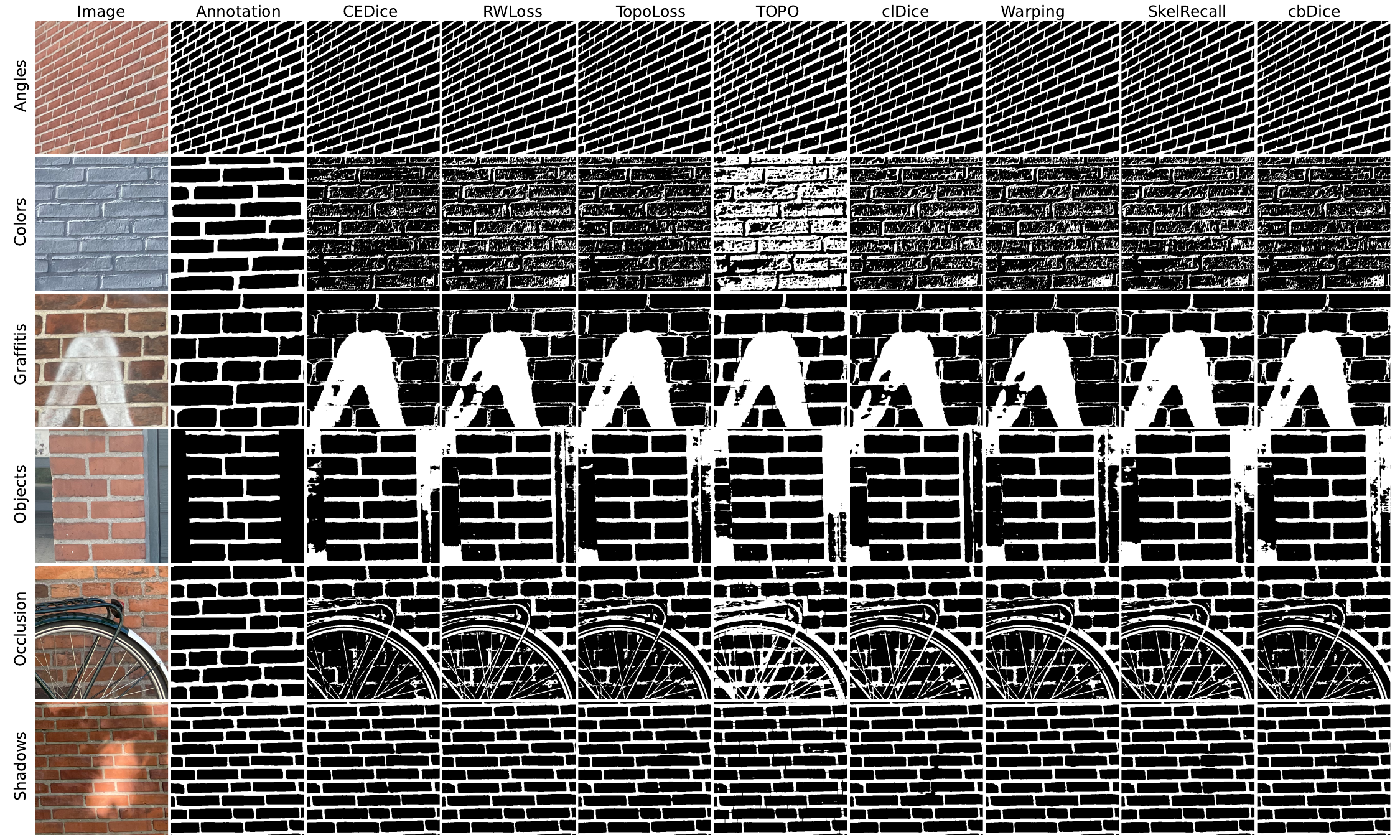}
   \caption{Segmentations with median performance obtained on the following training setup: Standard supervised learning, large training set, accurate labels.} \label{app_fig:ood_segmentation_results}
\end{figure*}

\Cref{app_fig:ood_segmentation_results} shows images and their corresponding labels, categorized by their out-of-distribution group, and the segmentations achieved by training on each loss function.

\section{Significance tests} \label{app_sec:significance_tests}
P-values were obtained by the paired permutation test comparing the Betti errors between methods. We considered results with p $<$ 0.05 to be statistically significant.

\begin{figure*}[h]
\centering
    \includegraphics[width=0.9\textwidth]{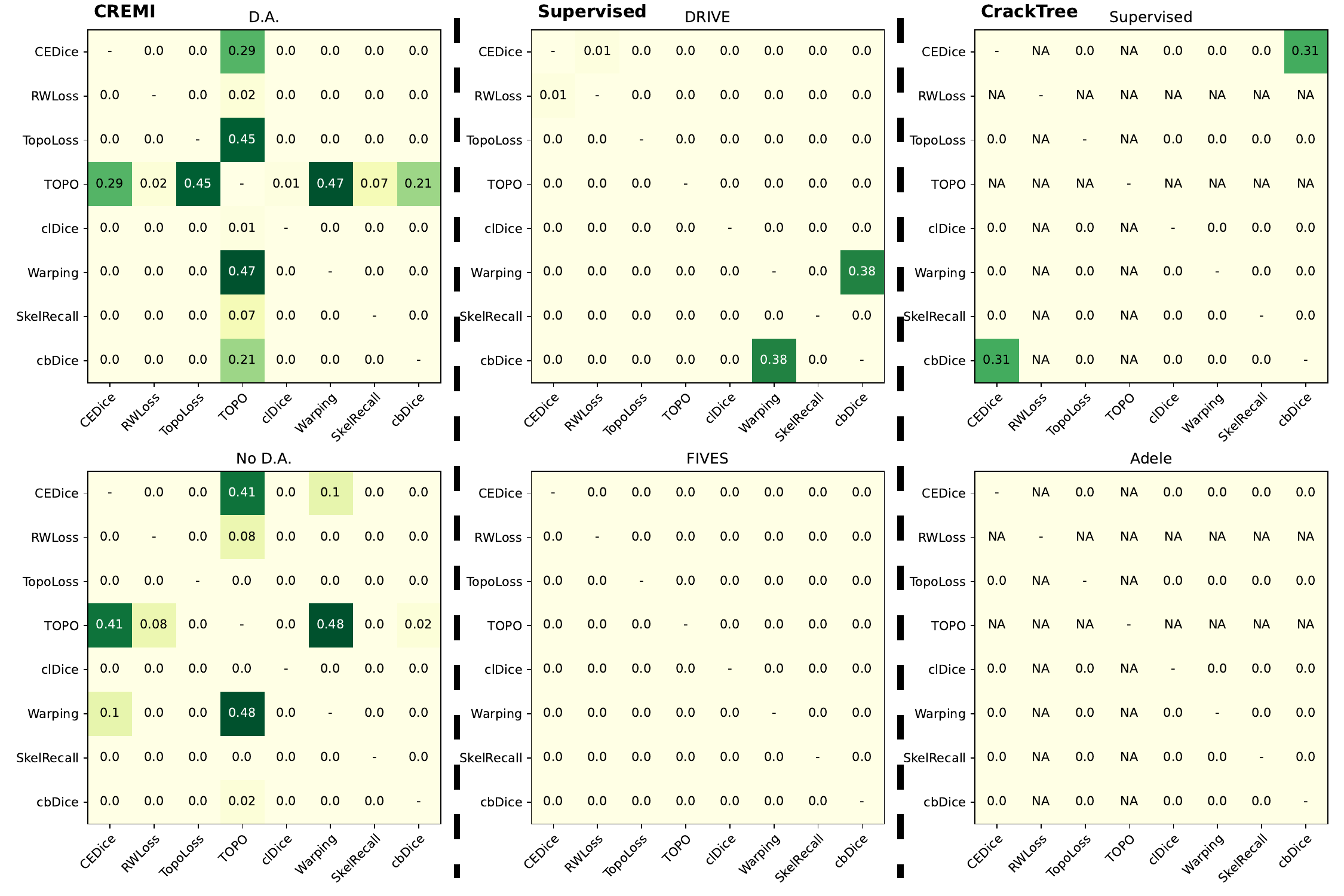}
   \caption{P-values corresponding to \Cref{table:preliminary} in \Cref{sec:preliminaryexperiments} ``Challenges and limitations in previous datasets". Training setup: (Left column) CREMI with and without DA, (middle column) DRIVE vs. FIVES datasets, (right column) CrackTree via standard supervised learning vs. via Adele.} \label{app_fig:pvalues_otherdatasets}
\end{figure*}

\begin{figure*}[h]
\centering
    \includegraphics[width=\textwidth]{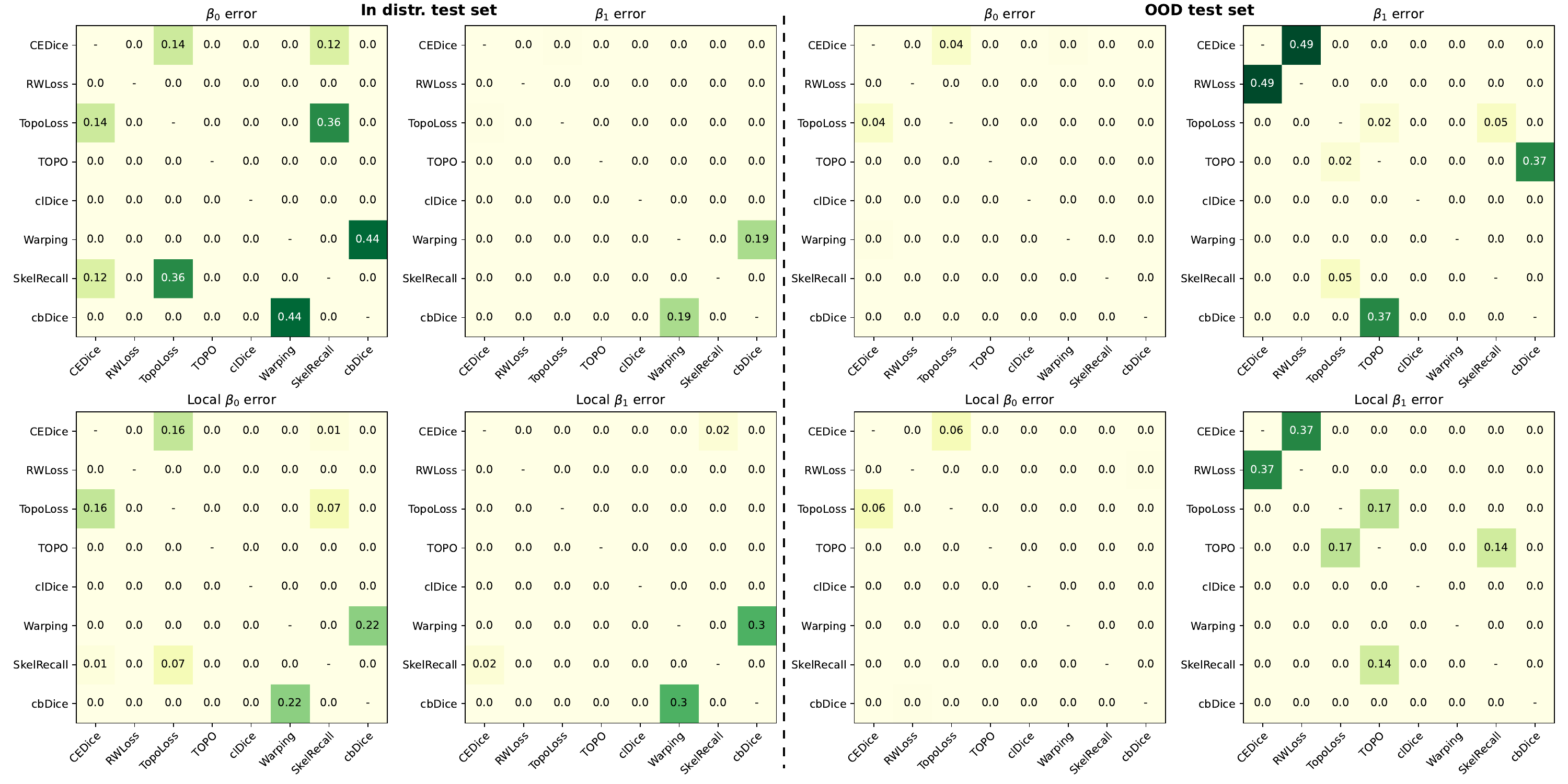}
   \caption{P-values corresponding to \Cref{table:standardbenchmark} in \Cref{sec:standardbenchmark} ``Benchmark on TopoMortar without challenges". Training setup: TopoMortar, standard supervised learning, large training set, accurate labels.} \label{app_fig:pvalues_topomortar_nochallenge}
\end{figure*}

\begin{figure*}[h]
\centering
    \includegraphics[width=\textwidth]{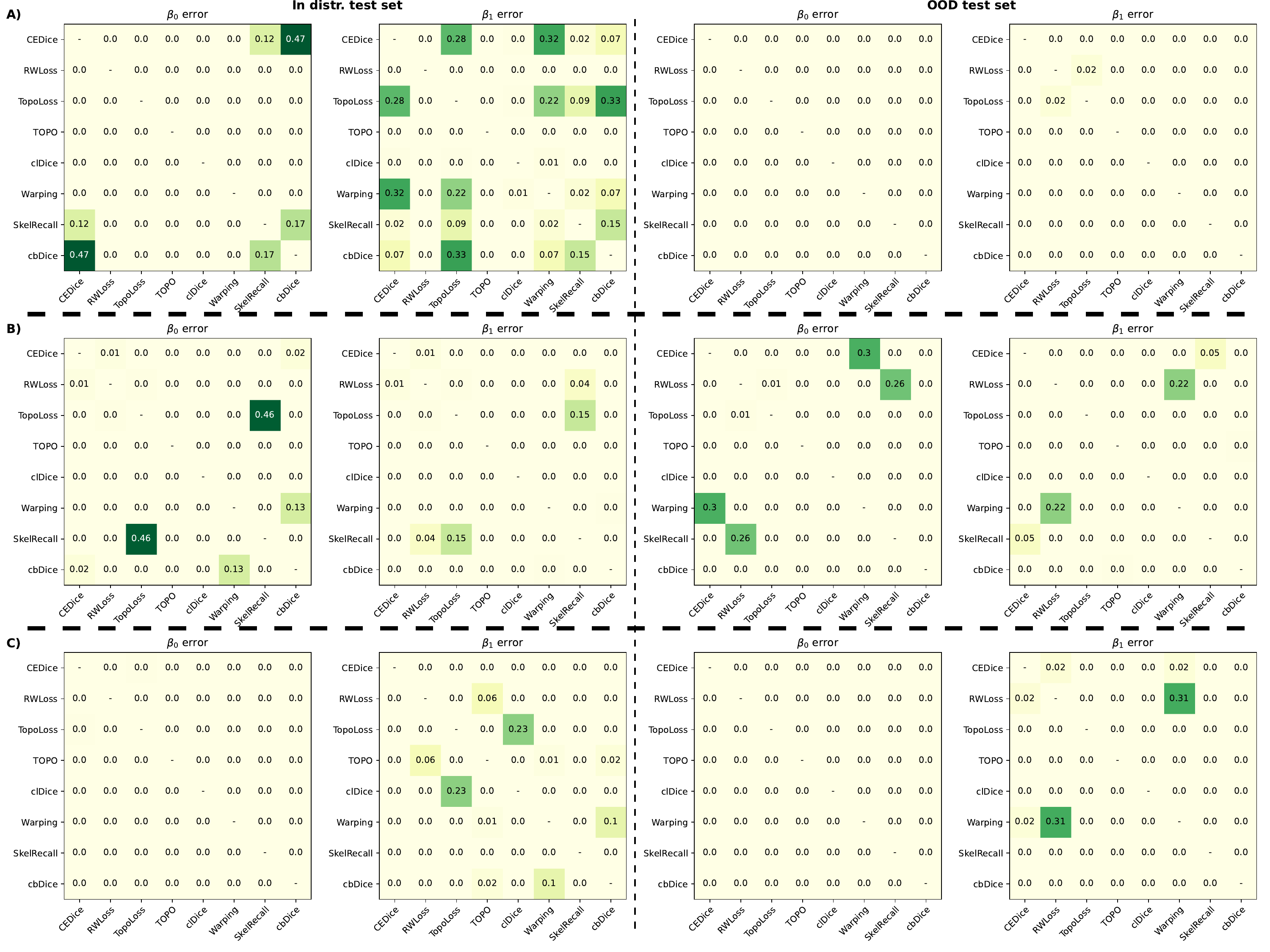}
   \caption{P-values corresponding to \Cref{table:challenges} in \Cref{sec:exprobust} ``Robustness to scarce training data, low-quality labels, and OOD images". A: Standard supervised learning, \textit{small training set}, accurate labels. B: Standard supervised, large training set, \textit{pseudo-labels}. C: Standard supervised learning, large training set, \textit{noisy labels}.} \label{app_fig:pvalues_topomortar_robustness}
\end{figure*}

\begin{figure*}[h]
\centering
    \includegraphics[width=\textwidth]{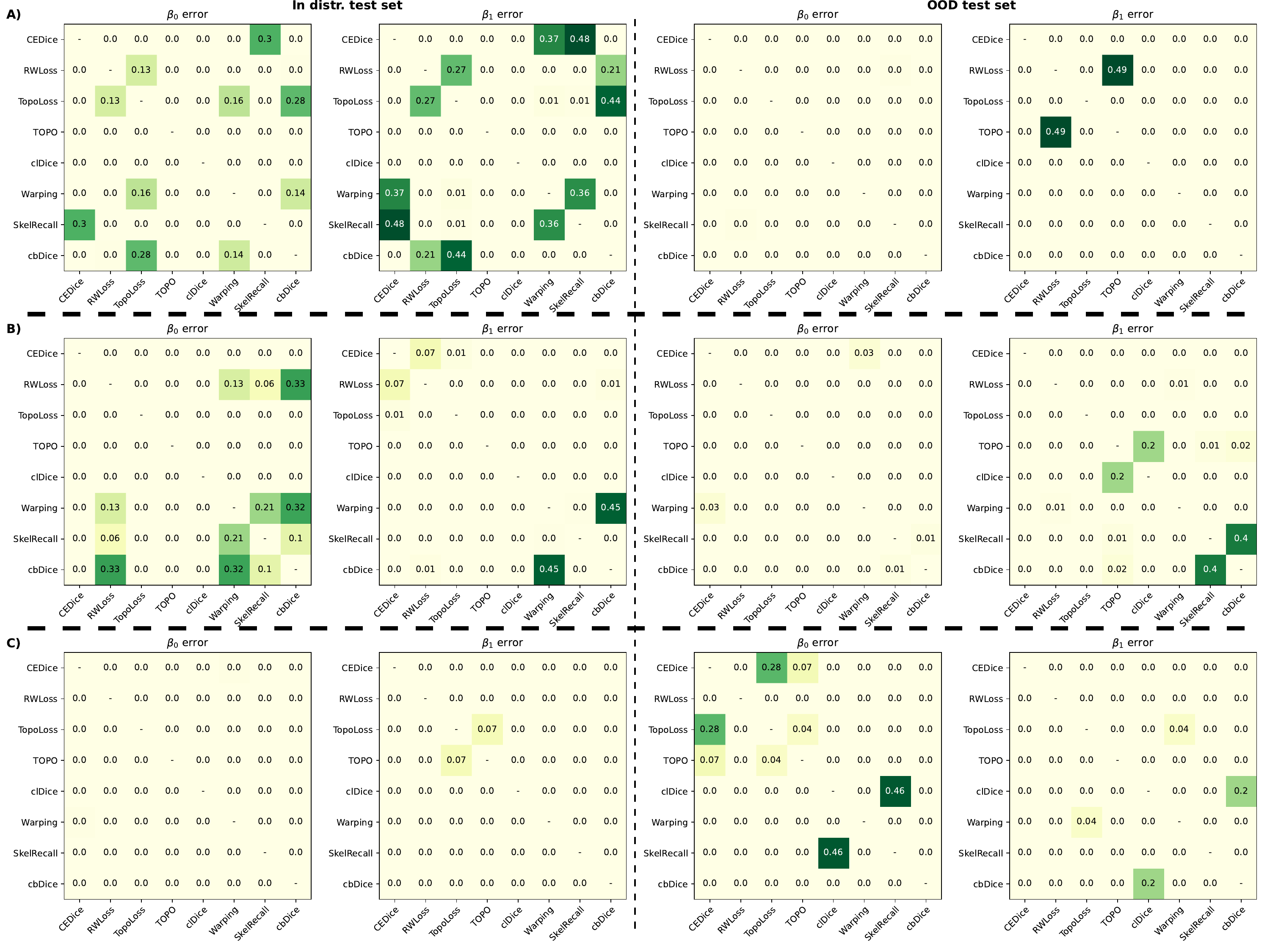}
   \caption{P-values corresponding to \Cref{table:challenges} in \Cref{sec:daself} ``Topology losses with data augmentation and self-distillation". A: Standard supervised learning, small training set, accurate labels, with the extra data augmentation \textit{RandHue}. B: \textit{Self-distillation}, large training set, \textit{pseudo-labels}. C: \textit{Self-distillation}, large training set, \textit{noisy labels}.} \label{app_fig:pvalues_topomortar_dadistil}
\end{figure*}

\begin{itemize}
    \item \Cref{app_fig:pvalues_otherdatasets}: p-values corresponding to \Cref{sec:preliminaryexperiments} ``Challenges and limitations in previous datasets" (\Cref{table:preliminary}).
    \item \Cref{app_fig:pvalues_topomortar_nochallenge}: p-values corresponding to \Cref{sec:standardbenchmark} ``Benchmark on TopoMortar without challenges" (\Cref{table:standardbenchmark}).
    \item \Cref{app_fig:pvalues_topomortar_robustness}: p-values corresponding to \Cref{sec:exprobust} ``Robustness to scarce training data, low-quality labels, and OOD images" (\Cref{table:challenges}).
    \item \Cref{app_fig:pvalues_topomortar_dadistil}: p-values corresponding to \Cref{sec:daself} ``Topology losses with data augmentation and self-distillation" (\Cref{table:challenges}).
\end{itemize}

\newpage
\clearpage

\section{Dice and HD95 measurements} \label{app_sec:dicehd95}
\Cref{app_table:dice_topomortar_robustness,app_table:dice_topomortar_dadistil} show the Dice coefficients and HD95 from \Cref{sec:exprobust} ``Robustness to scarce training data, low-quality labels, and OOD images" and \Cref{sec:daself} ``Topology losses with data augmentation and self-distillation", respectively.

\begin{table}[h]
\centering
{\footnotesize
\begin{tabular}{|ll|ll|ll|}
\hline
 &  & \multicolumn{2}{c|}{Dice}  & \multicolumn{2}{c|}{HD95}  \\
 & Test set $\rightarrow$ & ID & OOD & ID & OOD \\ \hline
\parbox[t]{1mm}{\multirow{8}{*}{\rotatebox[origin=c]{90}{\shortstack[c]{Small training set}}}} & CEDice & 0.9\std{0.0} & 0.55\std{0.0} & 2.06\std{0.11} & 39.57\std{1.11} \\
  & RWLoss & 0.9\std{0.0} & 0.57\std{0.01} & 2.04\std{0.06} & 39.46\std{1.35} \\
  & TopoLoss & 0.9\std{0.0} & 0.56\std{0.01} & 2.09\std{0.11} & 39.86\std{1.17} \\
  & TOPO & 0.86\std{0.0} & 0.59\std{0.01} & 2.64\std{0.22} & 40.54\std{0.58} \\
  & clDice & 0.9\std{0.0} & 0.56\std{0.01} & 2.04\std{0.09} & 39.83\std{1.88} \\
  & Warping & 0.9\std{0.0} & 0.56\std{0.01} & 2.06\std{0.07} & 39.56\std{1.26} \\
  & SkelRecall & 0.9\std{0.0} & 0.57\std{0.01} & 2.1\std{0.12} & 39.68\std{0.86} \\
  & cbDice & 0.9\std{0.0} & 0.56\std{0.01} & 2.08\std{0.11} & 39.62\std{0.92} \\
\hline
\parbox[t]{1mm}{\multirow{8}{*}{\rotatebox[origin=c]{90}{\shortstack[c]{Pseudo-labels}}}} & CEDice & 0.86\std{0.0} & 0.68\std{0.01} & 3.17\std{0.02} & 35.19\std{2.6} \\
  & RWLoss & 0.87\std{0.0} & 0.67\std{0.01} & 3.09\std{0.03} & 36.65\std{1.12} \\
  & TopoLoss & 0.86\std{0.0} & 0.68\std{0.01} & 3.18\std{0.02} & 36.33\std{2.51} \\
  & TOPO & 0.81\std{0.0} & 0.66\std{0.01} & 4.4\std{0.06} & 40.53\std{0.81} \\
  & clDice & 0.87\std{0.0} & 0.67\std{0.01} & 3.17\std{0.02} & 36.87\std{0.88} \\
  & Warping & 0.87\std{0.0} & 0.67\std{0.01} & 3.12\std{0.03} & 36.43\std{1.29} \\
  & SkelRecall & 0.86\std{0.0} & 0.7\std{0.01} & 3.29\std{0.03} & 34.91\std{1.37} \\
  & cbDice & 0.86\std{0.0} & 0.68\std{0.01} & 3.13\std{0.02} & 35.72\std{2.52} \\
\hline
\parbox[t]{1mm}{\multirow{8}{*}{\rotatebox[origin=c]{90}{\shortstack[c]{Noisy labels}}}} & CEDice & 0.63\std{0.0} & 0.33\std{0.01} & 3.84\std{0.02} & 40.17\std{2.02} \\
  & RWLoss & 0.66\std{0.0} & 0.31\std{0.01} & 3.57\std{0.01} & 36.35\std{1.51} \\
  & TopoLoss & 0.62\std{0.0} & 0.26\std{0.01} & 3.88\std{0.02} & 42.99\std{2.81} \\
  & TOPO & 0.84\std{0.01} & 0.58\std{0.01} & 2.65\std{0.19} & 39.27\std{0.84} \\
  & clDice & 0.69\std{0.0} & 0.46\std{0.01} & 3.58\std{0.01} & 38.9\std{1.46} \\
  & Warping & 0.62\std{0.0} & 0.35\std{0.0} & 3.85\std{0.01} & 37.61\std{0.98} \\
  & SkelRecall & 0.76\std{0.0} & 0.51\std{0.01} & 3.24\std{0.01} & 40.16\std{1.8} \\
  & cbDice & 0.63\std{0.0} & 0.31\std{0.01} & 3.82\std{0.01} & 39.26\std{2.18} \\
\hline
\end{tabular}
} \caption{Dice and HD95 measurements complementary to \Cref{table:challenges} in \Cref{sec:exprobust} ``Robustness to scarce training data, low-quality labels, and OOD images".}  \label{app_table:dice_topomortar_robustness}
\end{table}

\begin{table}[]
\centering
{\footnotesize
\begin{tabular}{|ll|ll|ll|}
\hline
 &  & \multicolumn{2}{c|}{Dice}  & \multicolumn{2}{c|}{HD95}  \\
 & Test set $\rightarrow$ & ID & OOD & ID & OOD \\ \hline
\parbox[t]{1mm}{\multirow{8}{*}{\rotatebox[origin=c]{90}{\shortstack[c]{D.A. (\textit{RandHue})}}}} & CEDice & 0.91\std{0.0} & 0.74\std{0.01} & 1.88\std{0.31} & 36.1\std{0.68} \\
  & RWLoss & 0.91\std{0.0} & 0.72\std{0.01} & 1.8\std{0.01} & 36.54\std{0.54} \\
  & TopoLoss & 0.91\std{0.0} & 0.74\std{0.01} & 1.9\std{0.52} & 36.19\std{0.52} \\
  & TOPO & 0.86\std{0.0} & 0.72\std{0.01} & 2.64\std{0.49} & 42.31\std{1.06} \\
  & clDice & 0.91\std{0.0} & 0.74\std{0.01} & 1.77\std{0.0} & 35.42\std{0.69} \\
  & Warping & 0.91\std{0.0} & 0.73\std{0.01} & 1.8\std{0.05} & 35.94\std{0.73} \\
  & SkelRecall & 0.91\std{0.0} & 0.75\std{0.01} & 1.91\std{0.28} & 37.83\std{0.72} \\
  & cbDice & 0.91\std{0.0} & 0.74\std{0.01} & 1.89\std{0.48} & 36.5\std{0.78} \\
\hline
\parbox[t]{1mm}{\multirow{8}{*}{\rotatebox[origin=c]{90}{\shortstack[c]{Pseudo + Self. dist.}}}} & CEDice & 0.87\std{0.0} & 0.69\std{0.01} & 3.11\std{0.04} & 36.34\std{2.24} \\
  & RWLoss & 0.87\std{0.0} & 0.67\std{0.01} & 3.11\std{0.03} & 36.09\std{1.04} \\
  & TopoLoss & 0.86\std{0.0} & 0.68\std{0.01} & 3.43\std{0.03} & 37.93\std{2.66} \\
  & TOPO & 0.76\std{0.0} & 0.63\std{0.01} & 5.56\std{0.37} & 39.74\std{0.97} \\
  & clDice & 0.88\std{0.0} & 0.63\std{0.01} & 3.24\std{0.08} & 37.82\std{0.56} \\
  & Warping & 0.88\std{0.0} & 0.66\std{0.01} & 3.01\std{0.03} & 36.48\std{1.48} \\
  & SkelRecall & 0.86\std{0.0} & 0.68\std{0.01} & 3.4\std{0.05} & 36.44\std{1.26} \\
  & cbDice & 0.87\std{0.0} & 0.68\std{0.01} & 3.06\std{0.05} & 37.29\std{2.19} \\
\hline
\parbox[t]{1mm}{\multirow{8}{*}{\rotatebox[origin=c]{90}{\shortstack[c]{Noisy + Self. dist.}}}} & CEDice & 0.65\std{0.0} & 0.35\std{0.01} & 3.72\std{0.02} & 42.12\std{2.15} \\
  & RWLoss & 0.66\std{0.0} & 0.3\std{0.01} & 3.62\std{0.11} & 36.97\std{1.81} \\
  & TopoLoss & 0.6\std{0.01} & 0.25\std{0.01} & 4.09\std{1.08} & 71.88\std{44.92} \\
  & TOPO & 0.83\std{0.02} & 0.54\std{0.02} & 3.52\std{0.85} & 41.72\std{1.06} \\
  & clDice & 0.78\std{0.0} & 0.49\std{0.01} & 2.98\std{0.01} & 42.41\std{3.58} \\
  & Warping & 0.67\std{0.0} & 0.37\std{0.0} & 3.56\std{0.02} & 42.07\std{2.47} \\
  & SkelRecall & 0.79\std{0.0} & 0.55\std{0.01} & 3.08\std{0.02} & 39.94\std{2.0} \\
  & cbDice & 0.69\std{0.0} & 0.35\std{0.0} & 3.44\std{0.02} & 42.83\std{1.75} \\
\hline
\end{tabular}
} \caption{Dice and HD95 measurements complementary to \Cref{table:challenges} in \Cref{sec:daself} ``Topology losses with data augmentation and self-distillation".} \label{app_table:dice_topomortar_dadistil}
\end{table}

\section{RandHue data augmentation} \label{app_sec:randhue_example}
\begin{figure*}
\centering
    \includegraphics[width=\textwidth]{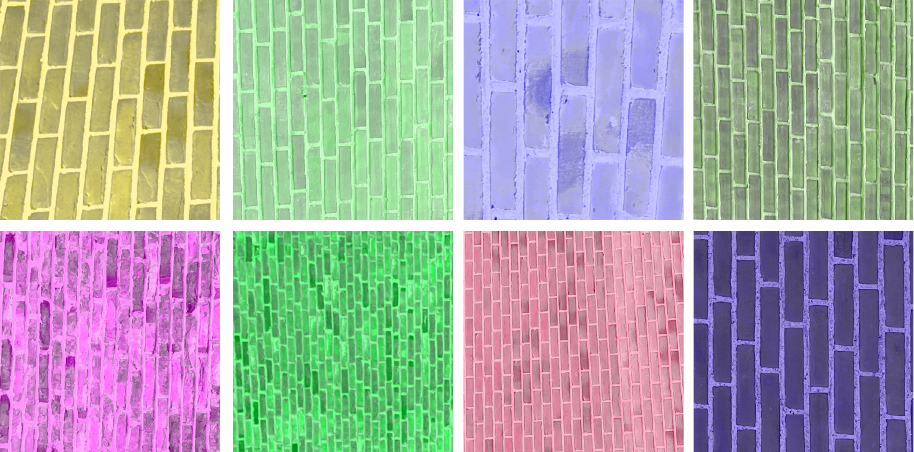}
   \caption{Examples of brick images augmented with RandHue} \label{app_fig:randhue_example}
\end{figure*}

\Cref{app_fig:randhue_example} illustrates representative examples of TopoMortar training images augmented with \textit{RandHue}.

\section{Baseline and RandHue results divided by OOD groups} \label{app_sec:topomortar_baselinerandhue_oodgroups}
\Cref{app_table:topomortar_baselinerandhue_oodgroups} shows the average $\beta_0$, $\beta_1$, Dice coefficient and HD95 across the different loss functions on the OOD test set, separating the measurements by OOD group.

\begin{table}
\centering
{\footnotesize
\begin{tabular}{|ll|ll|ll|}
\hline
 & OOD group & $\beta_0$ error & $\beta_1$ error & Dice & HD95 \\ \hline
\parbox[t]{1mm}{\multirow{6}{*}{\rotatebox[origin=c]{90}{\shortstack[c]{Baseline}}}} & Angles & 56.36\std{19.77} & 27.74\std{18.99} & 0.87\std{0.01} & 5.19\std{1.04} \\
  & Colors & 605.0\std{66.74} & 257.8\std{45.09} & 0.5\std{0.01} & 10.65\std{0.31} \\
  & Graffiti & 135.37\std{19.9} & 70.49\std{19.2} & 0.53\std{0.0} & 34.04\std{1.89} \\
  & Objects & 118.13\std{33.13} & 74.11\std{27.15} & 0.42\std{0.01} & 144.19\std{1.59} \\
  & Occlusion & 87.07\std{7.32} & 70.03\std{10.57} & 0.76\std{0.0} & 24.39\std{0.63} \\
  & Shadows & 68.44\std{16.89} & 35.85\std{15.53} & 0.83\std{0.01} & 8.62\std{1.34} \\
\hline
\parbox[t]{1mm}{\multirow{6}{*}{\rotatebox[origin=c]{90}{\shortstack[c]{RandHue}}}} & Angles & 34.69\std{9.29} & 17.92\std{5.8} & 0.86\std{0.01} & 5.84\std{0.37} \\
  & Colors & 77.25\std{12.71} & 47.12\std{8.44} & 0.79\std{0.01} & 5.42\std{0.3} \\
  & Graffiti & 95.24\std{10.78} & 80.97\std{11.27} & 0.6\std{0.01} & 31.46\std{0.71} \\
  & Objects & 87.99\std{12.69} & 75.34\std{13.45} & 0.53\std{0.01} & 145.83\std{1.66} \\
  & Occlusion & 74.1\std{7.91} & 73.46\std{8.74} & 0.78\std{0.0} & 23.98\std{0.26} \\
  & Shadows & 38.96\std{12.33} & 30.86\std{10.42} & 0.85\std{0.01} & 10.08\std{0.98} \\
\hline
\end{tabular}
} \caption{Average measurements across loss functions per OOD group. Baseline: Corresponding to \Cref{sec:standardbenchmark} ``Benchmark on TopoMortar without challenges" (\Cref{table:standardbenchmark}). RandHue: Corresponding to \Cref{sec:daself} ``Topology losses with data augmentation and self-distillation" (\Cref{table:challenges}, \textit{RandHue})} \label{app_table:topomortar_baselinerandhue_oodgroups}
\end{table}

\section{Computational resources} \label{app:trainingresources}
\Cref{table:trainingresources} lists the computational resources (GPU memory, training time) of the different loss functions.

\begin{table}[h]
\centering
{\footnotesize
\begin{tabular}{|l|ll|}
\hline
Loss & GPU (GiB) & Time (h.) \\
\hline
CEDice      &   5.6  &     2.3  \\
RWLoss     &   5.6    &   14.5   \\
TopoLoss       &    5.6   &  49.6    \\
TOPO           &   5.9    &    28.8  \\
clDice         &    16.2   &   3.1   \\
Warping        &   5.6    &   152.1   \\
SkelRecall &   5.6    &   7.1   \\
cbDice         &   7.8    &   58.3   \\
\hline
\end{tabular}
}
\caption{Computational requirements for training a nnUNet on TopoMortar for 12,000 iterations with a batch size of 10. Hardware: Intel Xeon Gold 6126, NVIDIA Tesla V100 (32GB). } \label{table:trainingresources}
\end{table}

\section{High correlation between topology accuracy on TopoMortar and other datasets} \label{app_sec:highcorrelation}

TopoMortar is designed as a dataset that permits to control for dataset task confounding variables by fixing a task (segmenting mortar in red brick wall images) in order to study the individual effect on topology accuracy of four dataset challenges: small training set, noisy labels, pseudo-labels, and OOD test-set images.
This, ultimately, allows to elucidate the context in which topology-focused image segmentation methods, such as topology loss functions, are advantageous.
Importantly, although TopoMortar task is on mortar segmentation, our results are extrapolable to other datasets, which demonstrates the \textbf{generalizability of our conclusions} to biology and non-biology datasets, and to structures with different topology.

\Cref{sup_table:highcorr} shows the Pearson correlation between the topology accuracy obtained by topology loss functions in TopoMortar and the topology accuracy obtained in CREMI, DRIVE, FIVES, and CrackTree datasets.
The high correlations demonstrate 1) that TopoMortar can represent dataset challenges (\Cref{sup_table:highcorr}, first three rows), and 2) that TopoMortar can represent the results obtained after tackling dataset challenges (\Cref{sup_table:highcorr}, last two rows).

\begin{table*}[h]
\centering
{\footnotesize
\begin{tabular}{|l|l|l|}
\hline
Setting 1 & Setting 2 & Corr. \\
\hline
CREMI, Data augmentation, $\beta_1$  &   TopoMortar, Pseudo labels, In distribution, $\beta_1$  &  0.933  \\
DRIVE, $\beta_0$ & TopoMortar, Small training set, In distribution, $\beta_0$ & 0.905 \\
CrackTree, Supervised, $\beta_0$ & TopoMortar, Noisy labels, In distribution, $\beta_0$ & 0.643 \\
FIVES, $\beta_0$ & TopoMortar, Large training set, In distribution, $\beta_0$ & 0.782 \\
CrackTree, Adele, $\beta_0$ & TopoMortar, Noisy labels, Self-distillation, $\beta_0$ & 0.920 \\
\hline
\end{tabular}
}
\caption{Pearson correlation between experimental settings using existing datasets and their corresponding representation in TopoMortar.} \label{sup_table:highcorr}
\end{table*}

\end{document}